\documentclass[10pt,journal,compsoc]{IEEEtran}
\ifCLASSOPTIONcompsoc
  \usepackage[nocompress]{cite}
\else
  \usepackage{cite}
\fi
\ifCLASSINFOpdf
   \usepackage[pdftex]{graphicx}
\else
   \usepackage[dvips]{graphicx}
\fi
\usepackage{amsmath}
\usepackage{algorithmic}

\usepackage{array}

\ifCLASSOPTIONcompsoc
  \usepackage[caption=false,font=footnotesize,labelfont=sf,textfont=sf]{subfig}
\else
  \usepackage[caption=false,font=footnotesize]{subfig}
\fi
\usepackage{fixltx2e}

\usepackage{stfloats}
\usepackage{url}

\usepackage{color}
\usepackage{bm}

\usepackage{booktabs}
\usepackage{multirow}
\usepackage{makecell}
\usepackage{amsthm,amssymb,amsfonts}
\usepackage{diagbox}
\usepackage{footnote}
\begin{document}
%
\title{Recent Advances in Open Set Recognition: A Survey}
%
%
%
%

\author{Chuanxing~Geng, Sheng-Jun Huang and Songcan~Chen
\IEEEcompsocitemizethanks{\IEEEcompsocthanksitem The authors are with College of Computer Science and Technology, Nanjing University of Aeronautics and Astronautics of (NUAA), Nanjing, 211106, China.\protect\\
E-mail: \{gengchuanxing; huangsj; s.chen\}@nuaa.edu.cn.
\IEEEcompsocthanksitem Corresponding author is Songcan Chen.}
\thanks{Manuscript received April 19, XXXX; revised August 26, XXXX.}}

%
%

\markboth{Journal of \LaTeX\ Class Files,~Vol.~14, No.~8, August~2015}%
{Shell \MakeLowercase{\textit{et al.}}: Bare Demo of IEEEtran.cls for Computer Society Journals}
%



\IEEEtitleabstractindextext{%
\begin{abstract}
In real-world recognition/classification tasks, limited by various objective factors, it is usually difficult to collect training samples to exhaust all classes when training a recognizer or classifier. A more realistic scenario is open set recognition (OSR), where incomplete knowledge of the world exists at training time, and unknown classes can be submitted to an algorithm during testing, requiring the classifiers to not only accurately classify the seen classes, but also effectively deal with unseen ones. This paper provides a comprehensive survey of existing open set recognition techniques covering various aspects ranging from related definitions, representations of models, datasets, evaluation criteria, and algorithm comparisons. Furthermore, we briefly analyze the relationships between OSR and its related tasks including zero-shot, one-shot (few-shot) recognition/learning techniques, classification with reject option, and so forth. Additionally, we also review the open world recognition which can be seen as a natural extension of OSR. Importantly, we highlight the limitations of existing approaches and point out some promising subsequent research directions in this field.
\end{abstract}

}

\maketitle

\IEEEdisplaynontitleabstractindextext

%
\IEEEpeerreviewmaketitle

\IEEEraisesectionheading{\section{Introduction}\label{sec:introduction}}
\IEEEPARstart{U}{nder} a common closed set (or static environment) assumption: the training and testing data are drawn from the same label and feature spaces, the traditional recognition/classification algorithms have already achieved significant success in a variety of machine learning (ML) tasks. However, a more realistic scenario is usually open and non-stationary such as driverless, fault/medical diagnosis, etc., where unseen situations can emerge unexpectedly, which drastically weakens the robustness of these existing methods. To meet this challenge, several related research topics have been explored including lifelong learning \cite{Thrun1995Lifelong,Pentina2014A}, transfer learning \cite{Pan2010A,Weiss2016A,shao2015transfer}, domain adaptation \cite{Patel2015Visual,yamada2014domain}, zero-shot \cite{Palatucci2009Zero,Lampert2009Learning,fu2018recent}, one-shot (few-shot) \cite{Li2006One,Lake2013One,Li2008A,Amit2007Uncovering,Torralba2010Using,fu2013learning,vinyals2016matching,Snell2017Prototypical,Bertinetto2016Learning,Chen2018Semantic} recognition/learning, open set recognition/classification \cite{scheirer2012toward,Scheirer2014Probability,Jain2014Multi}, and so forth.

Based on Donald Rumsfeld's famous ``There are known knowns'' statement \cite{Ross2010Known}, we further expand the basic recognition categories of classes asserted by \cite{Scheirer2014Probability}, where we restate that recognition should consider four basic categories of classes as follows:
\begin{enumerate}[\IEEEsetlabelwidth{8)}]
\item \emph{known known classes} (KKCs), i.e., the classes with distinctly labeled positive training samples (also serving as negative samples for other KKCs), and even have the corresponding side-information like semantic/attribute information, etc;
\item \emph{known unknown classes} (KUCs), i.e., labeled negative samples, not necessarily grouped into meaningful classes, such as the background classes\cite{dhamija2018reducing}, the universum classes\cite{weston2006inference}\footnote{The universum classes\cite{weston2006inference} usually denotes the samples that do not belong to either class of interest for the specific learning problem.}, etc;
\item \emph{$\text{unknown known classes}^{\ast}$\footnote{$\ast$ represents the expanded basic recognition class by ourselves.}} (UKCs), i.e., classes with no available samples in training, but available side-information (e.g., semantic/attribute information) of them during training;
\item \emph{unknown unknown classes} (UUCs), i.e., classes without any information regarding them during training: not only unseen but also having not side-information (e.g., semantic/attribute information, etc.) during training.
\end{enumerate}

\begin{figure}[]
  \centering
  \includegraphics[width=0.4\textwidth]{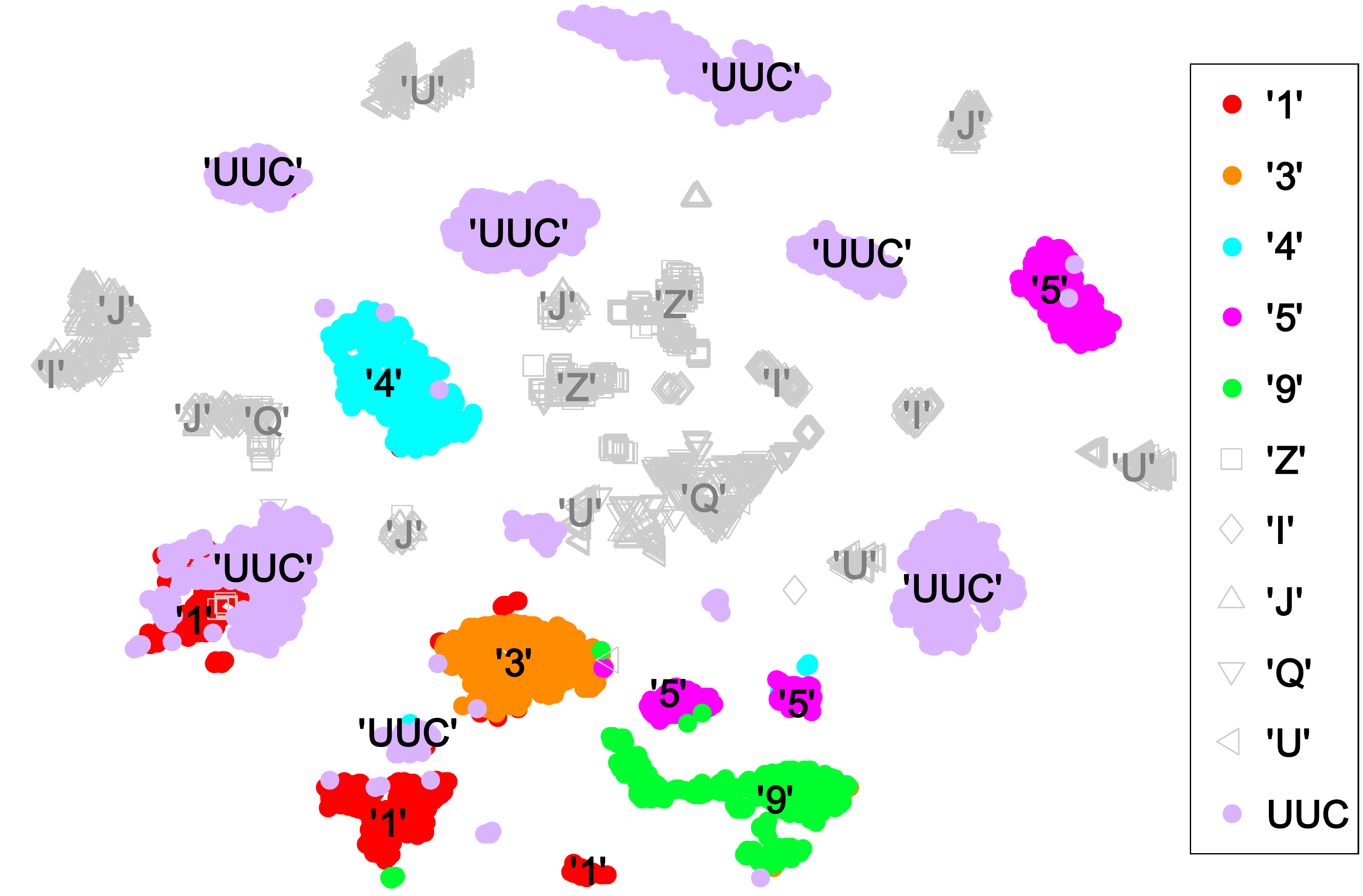}
  \caption{ An example of visualizing KKCs, KUCs, and UUCs from the real data distribution using t-SNE. Here, '1','3','4','5','9' are randomly selected from PENDIGITS as KKCs, while the remaining classes in it as UUCs. 'Z','I','J','Q','U' are randomly selected from LETTER as KUCs. This visualization also indicates that the distribution of one class may consist of multiple subclass/subclusters, e.g., class '1', '5', 'U', etc.}
  \label{fig_sim}
\end{figure}

\begin{figure*}[!t]
\centering
\subfloat[Distribution of the original data set.]{\includegraphics[width=2.2in]{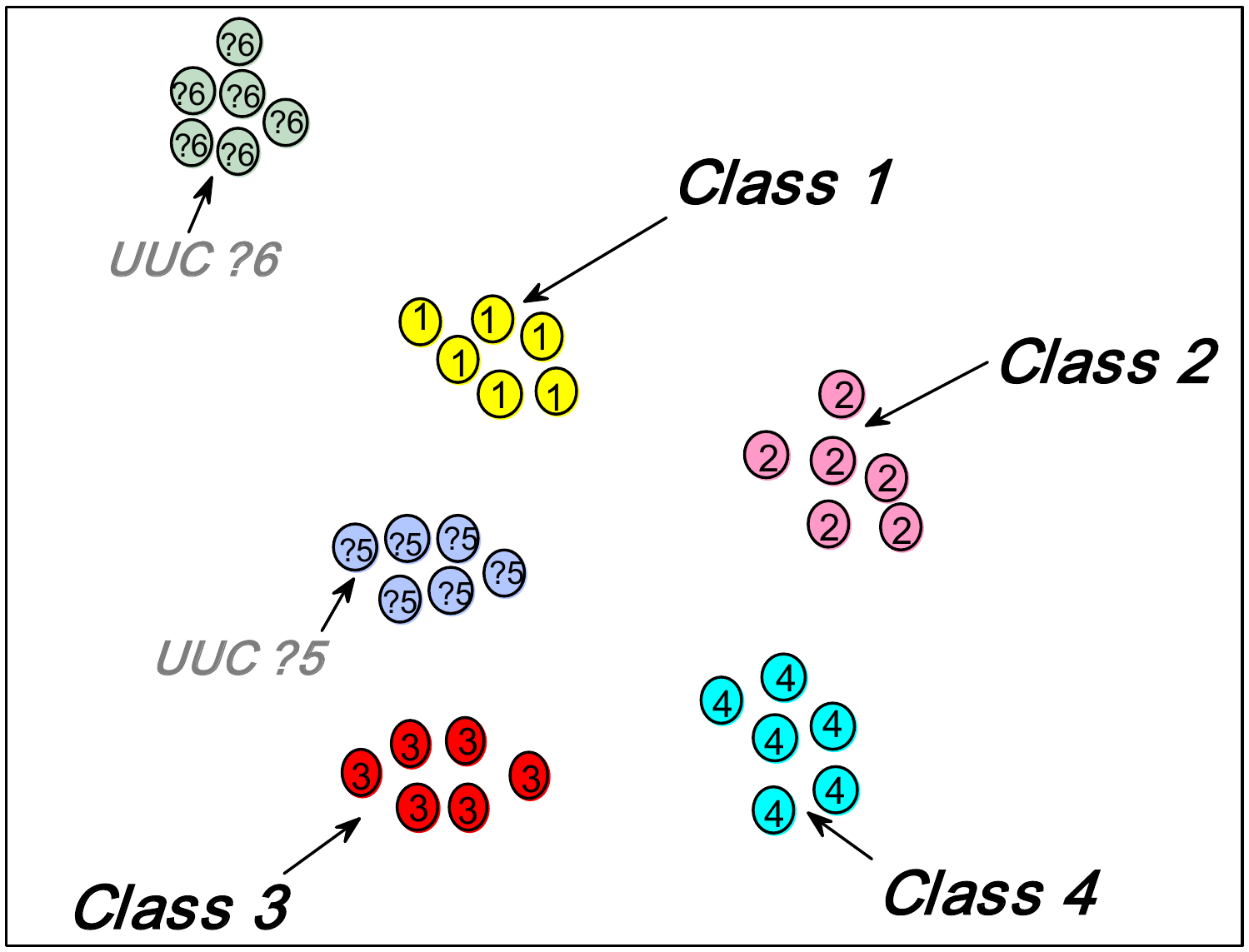}%
\label{fig_first_case}}
\hfil
\subfloat[Traditional recognition/classification problem.]{\includegraphics[width=2.25in]{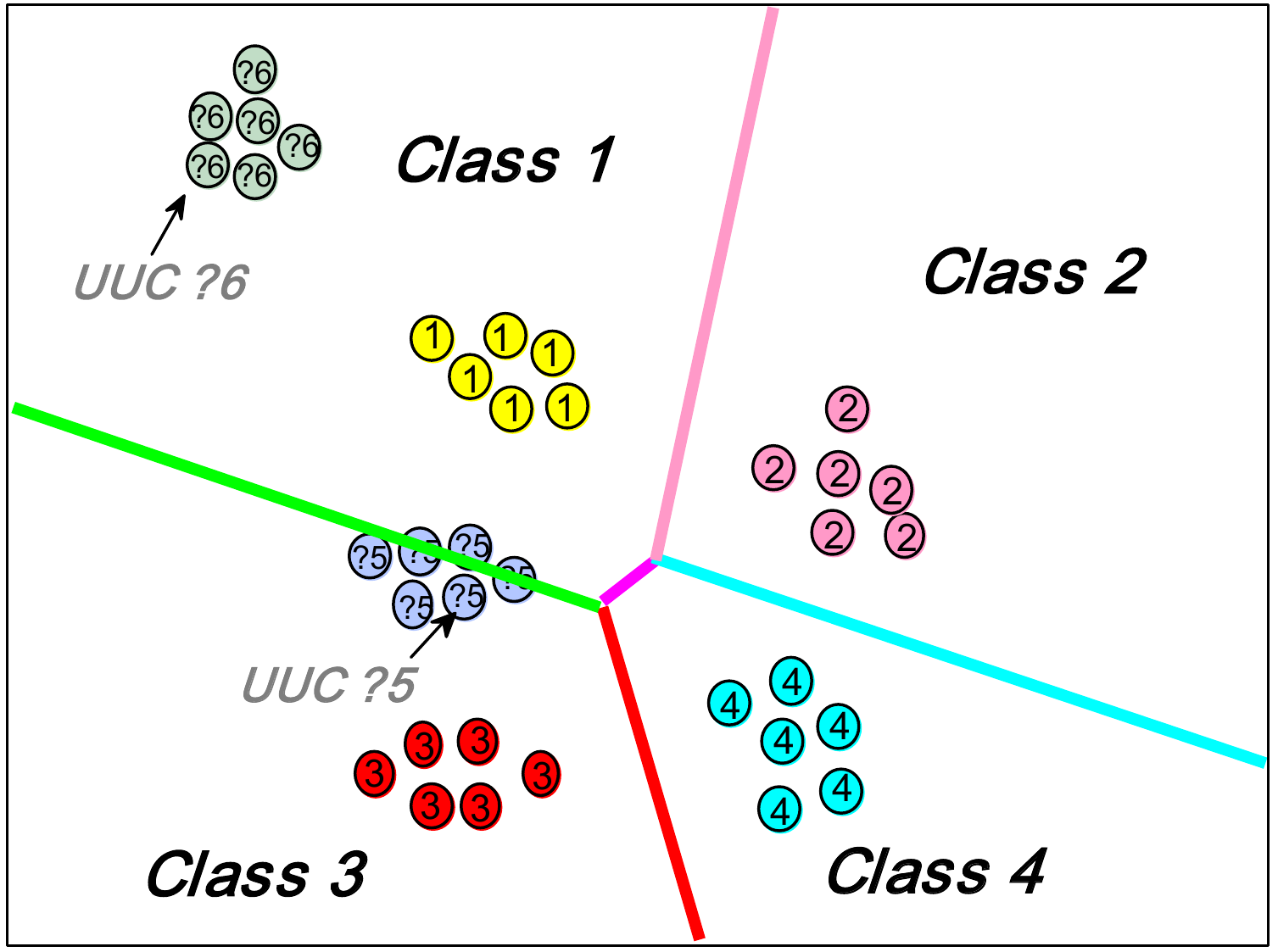}%
\label{fig_second_case}}
\hfil
\subfloat[Open set recognition/classification problem.]{\includegraphics[width=2.21in]{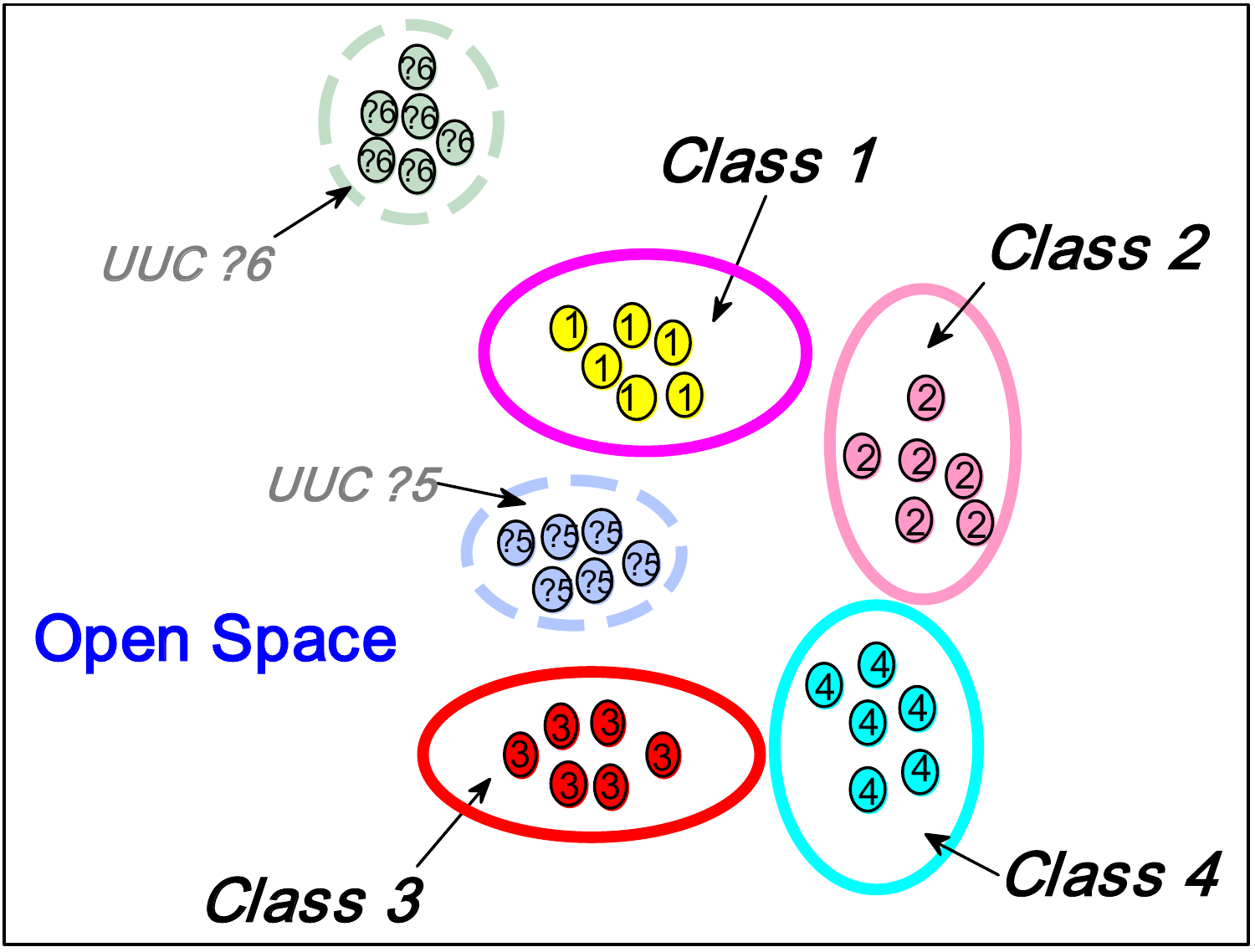}%
\label{fig_third_case}}
\caption{Comparison of between traditional classification and open set recognition. Fig. 2(a) denotes the distribution of original dataset including KKCs 1,2,3,4 and UUCs ?5,?6, where KKCs appear during training and testing while UUCs may appear or not during testing. Fig. 2(b) shows the decision boundary of each class obtained by traditional classification methods, and it will obviously misclassify when UUCs ?5,?6 appear during testing. Fig. 2(c) describes open set recognition, where the decision boundaries limit the scope of KKCs 1,2,3,4, reserving space for UUCs ?5,?6. Via these decision boundaries, the samples from some UUCs are labeled as "unknown" or rejected rather than misclassified as KKCs.}
\label{fig_sim}
\end{figure*}

Fig. 1 gives an example of visualizing KKCs, KUCs, and UUCs from the real data distribution using t-SNE\cite{maaten2008visualizing}. Note that since the main difference between UKCs and UUCs lies in whether their side-information is available or not, we here only visualize UUCs. Traditional classification only considers KKCs, while including KUCs will result in models with an explicit ''other class,'' or a detector trained with unclassified negatives \cite{Scheirer2014Probability}. Unlike traditional classification, zero-shot learning (ZSL) focuses more on the recognition of UKCs. As the saying goes: prediction is impossible without any assumptions about how the past relates to the future. ZSL leverages semantic information shared between KKCs and UKCs to implement such a recognition \cite{Palatucci2009Zero,Lampert2009Learning}. In fact, assuming the test samples only from UKCs is rather restrictive and impractical, since we usually know nothing about them either from KKCs or UKCs. On the other hand, the object frequencies in natural world follow long-tailed distributions \cite{salakhutdinov2011learning,zhu2014capturing}, meaning that KKCs are more common than UKCs. Therefore, some researchers have begun to pay attention to the more generalized ZSL (G-ZSL) \cite{socher2013zero,chao2016empirical,xian2018zero,felix2018multi}, where the testing samples come from both KKCs and UKCs. As a closely-related problem to ZSL, one/few-shot learning (FSL) can be seen as natural extensions of zero-shot learning when a limited number of UKCs' samples during training are available \cite{Li2006One,Lake2013One,Li2008A,Amit2007Uncovering,Torralba2010Using,fu2013learning,vinyals2016matching,Snell2017Prototypical,Bertinetto2016Learning,Chen2018Semantic}. Similar to G-ZSL, a more realistic setting for FSL considering both KKCs and UKCs in testing, i.e., generalized FSL (G-FSL), is also becoming more popular \cite{gidaris2018dynamic}.
Compared to (G-)ZSL and (G-)FSL, open set recognition (OSR)\cite{scheirer2012toward,Jain2014Multi,Scheirer2014Probability} probably faces a more serious challenge due to the fact that only KKCs are available without any other side-information like attributes or  a limited number of samples from UUCs.

Open set recognition \cite{scheirer2012toward} describes such a scenario where new classes (UUCs) unseen in training appear in testing, and requires the classifiers to not only accurately classify KKCs but also effectively deal with UUCs. Therefore, the classifiers need to have a corresponding reject option when a testing sample comes from some UUC. Fig. 2 gives a comparative demonstration of traditional classification and OSR problems. It should be noted that there have been already a variety of works in the literature regarding classification with reject option \cite{chow1970optimum,bartlett2008classification,tax2008growing,fischer2016optimal,fumera2002support,grandvalet2009support,herbei2006classification,yuan2010classification,zhang2006ro,Wegkamp2007Lasso,Geifman2017Selective}. Although related in some sense, this task should not be confused with open set recognition since it still works under the closed set assumption, while the corresponding classifier rejects to recognize an input sample due to its low confidence, avoiding classifying a sample of one class as a member of another one.

In addition, the one-class classifier \cite{scholkopf2001estimating,manevitz2001one,tax2004support,khan2009survey,Jin2004Face,Wu2009A,Cevikalp2012Efficient,pimentel2014review} usually used for anomaly detection seems suitable for OSR problem, in which the empirical distribution of training data is modeled such that it can be separated from the surrounding open space (the space far from known/training data) in all directions of the feature space. Popular approaches for one-class classification include one-class SVM \cite{scholkopf2001estimating} and support vector data description (SVDD) \cite{tax2004support,gornitz2018support}, where one-class SVM separates the training samples from the origin of the feature space with a maximum margin, while SVDD encloses the training data with a hypersphere of minimum volume. Note that treating multiple KKCs as a single one in the one-class setup obviously ignores the discriminative information among these KKCs, leading to poor performance \cite{bodesheim2013kernel,Jain2014Multi}. Even if each KKC is modeled by an individual one-class classifier as proposed in \cite{tax2008growing}, the novelty detection performance is rather low \cite{bodesheim2013kernel}. Therefore, it is necessary to rebuild effective classifiers specifically for OSR problem, especially for multiclass OSR problem.

\begin{table*}[]
\caption{Differences between Open Set Recognition and its related tasks}
\centering
\tabcolsep 1.6mm
\renewcommand\arraystretch{1.5}
\begin{tabular}{|c|c|c|c|}
\hline
\diagbox{\textbf{TASK}}{\textbf{SETTING}}                                                                        & \textbf{TRAINING}                                                                                                   & \textbf{TESTING}                                                                                   & \textbf{GOAL}                                                                                                           \\ \hline
\textbf{Traditional Classification}    & Known known classes  & Known known classes  & Classifying known known classes  \\ \hline
\textbf{Classification with Reject Option }                                                     & Known known classes                                                                                        & Known known classes                                                                       & \begin{tabular}[c]{@{}c@{}}Classifying known known classes \&\\ rejecting samples of low confidence\end{tabular}     \\ \hline
\begin{tabular}[c]{@{}c@{}}\textbf{One-class Classification}\\ (\textbf{Anomaly Detection})\end{tabular} & \begin{tabular}[c]{@{}c@{}}Known known classes \& few \\ or none outliers from KUCs\end{tabular}                     & \begin{tabular}[c]{@{}c@{}}Known known classes \& \\ few or none outliers\end{tabular}    & Detecting outliers                                                                                             \\ \hline
\begin{tabular}[c]{@{}c@{}}\textbf{One/Few-shot Learning}\end{tabular}        & \begin{tabular}[c]{@{}c@{}}Known known classes \& a limited\\   number of UKCs' samples\end{tabular}            & Unknown known classes           & Identifying unknown known classes                                                                              \\ \hline
\textbf{Generalized Few-shot Learning} & \begin{tabular}[c]{@{}c@{}}Known known classes \& a limited \\  number of UKCs' samples\end{tabular}     & \begin{tabular}[c]{@{}c@{}}Known known classes \& \\ unknown known classes\end{tabular}   & \begin{tabular}[c]{@{}c@{}}Identifying known known classes \& \\ unknown known classes\end{tabular}            \\ \hline
\textbf{Zero-shot Learning}                                                                     & \begin{tabular}[c]{@{}c@{}}Known known classes \& \\$\text{side-information}^1$\end{tabular}                         & Unknown known classes                                                                     & Identifying unknown known classes                                                                              \\ \hline
\textbf{Generalized Zero-shot Learning}                        & \begin{tabular}[c]{@{}c@{}}Known known classes \& \\$\text{side-information}^1$\end{tabular}     & \begin{tabular}[c]{@{}c@{}}Known known classes \& \\ unknown known classes\end{tabular}   & \begin{tabular}[c]{@{}c@{}}Identifying known known classes \& \\ unknown known classes\end{tabular}            \\ \hline
\textbf{Open Set Recognition }                                                                  & Known known classes                                                                                        & \begin{tabular}[c]{@{}c@{}}Known known classes \& \\ unknown unknown classes\end{tabular} & \begin{tabular}[c]{@{}c@{}}Identifying known known classes \&\\ rejecting unknown unknown classes\end{tabular}   \\ \hline\hline
\textbf{Generalized Open Set Recognition }                                                      & \begin{tabular}[c]{@{}c@{}}Known known classes \& \\ $\text{side-information}^2$\end{tabular} & \begin{tabular}[c]{@{}c@{}}Known known classes \& \\ Unknown unknown classes\end{tabular}   & \begin{tabular}[c]{@{}c@{}}Identifying known known classes \&\\ ’cognizing’ unknown unknown classes\end{tabular} \\ \hline
\end{tabular}

\emph{Note that the unknown known classes in one-shot learning usually do not have any side-information, such as semantic information, etc. The $\text{side-information}^1$ in ZSL and G-ZSL denotes the semantic information from both KKCs and UKCs, while the $\text{side-information}^2$ here denotes the available semantic information only from KKCs. As part of this information usually spans between KKCs and UUCs, we hope to use it to further 'cognize' UUCs instead of simply rejecting them.}
\end{table*}

As a summary, Table 1 lists the differences between open set recognition and its related tasks mentioned above. In fact, OSR has been studied under a number of frameworks, assumptions, and names \cite{Phillips2005Evaluation,Li2005Open,wu2007novel,wang2009support,Heflin2012Detecting,Pritsos2013Open}. In a study on evaluation methods for face recognition, Phillips et al. \cite{Phillips2005Evaluation} proposed a typical framework for open set identity recognition, while Li and Wechsler \cite{Li2005Open} again viewed open set face recognition from an evaluation perspective and proposed Open Set TCM-kNN (Transduction Confidence Machine-k Nearest Neighbors) method. It is Scheirer et al. \cite{scheirer2012toward} that first formalized the open set recognition problem and proposed a preliminary solution---1-vs-Set machine, which incorporates an open space risk term in modeling to account for the space beyond the reasonable support of KKCs. Afterwards, open set recognition attracted widespread attention. Note that OSR has been mentioned in the recent survey on ZSL \cite{fu2018recent}, however, it has not been extensively discussed. Unlike \cite{fu2018recent}, we here provide a comprehensive review regarding OSR.

The rest of this paper is organized as follows. In the next three sections, we first give the basic notation and related definitions (Section 2). Then we categorize the existing OSR technologies from the modeling perspective, and for each category, we review different approaches, given in Table 2 in detail (Section 3). Lastly, we review the open world recognition (OWR) which can be seen as a natural extension of OSR in Section 4. Furthermore, Section 5 reports the commonly used datasets, evaluation criteria, and algorithm comparisons, while Section 6 highlights the limitations of existing approaches and points out some promising research directions in this field. Finally, Section 7 gives a conclusion.

\section{Basic Notation and Related Definition}
This part briefly reviews the formalized OSR problem described in \cite{scheirer2012toward}. As discussed in \cite{scheirer2012toward}, the space far from known data (including KKCs and KUCs) is usually considered as \emph{open space} $\mathcal{O}$. So labeling any sample in this space as an arbitrary KKC inevitably incurs risk, which is called \emph{open space risk} $R_\mathcal{O}$. As UUCs are agnostic in training, it is often difficult to quantitatively analyze open space risk. Alternatively, \cite{scheirer2012toward} gives a qualitative description for $R_\mathcal{O}$, where it is formalized as the relative measure of open space $\mathcal{O}$ compared to the overall measure space $S_o$:
\begin{equation}
R_\mathcal{O}(f) = \frac{\int_\mathcal{O}f(x)dx}{\int_{S_o}f(x)dx},
\end{equation}
where $f$ denotes the measurable recognition function. $f(x)=1$ indicates that some class in KKCs is recognized, otherwise $f(x)=0$.
Under such a formalization, the more we label samples in open space as KKCs, the greater $R_\mathcal{O}$ is.

\begin{table*}[]
\caption{Different Categories of Models for Open Set Recognition}
\tabcolsep 1.4mm
\renewcommand\arraystretch{1.4}
\begin{center}
\begin{tabular}{|l|l|l|}
\hline
\multicolumn{2}{|c|}{Different Categories of OSR methods}             & \multicolumn{1}{c|}{Papers}  \\ \hline
\multirow{2}{*}{Discriminative model} & Traditional ML-based          & \cite{scheirer2012toward,Cevikalp2013Face,Cevikalp2017Best,Scheirer2014Probability,Jain2014Multi,Scherreik2016Open,Cevikalp2017Polyhedral,
Cevikalp2017Fast,Zhang2017Sparse,Bendale2015Towards,J2017Nearest,Rudd2018The,Vignotto2018Extreme,Fei2016Breaking,Vareto2018Towards,Neira2018Data}   \\ \cline{2-3}
                                      & Deep Neural Network-based     & \cite{Bendale2015Towards1,Rozsa2017Adversarial,Hassen2018Learning,prakhya2017open,Shu2017DOC,kardan2017mitigating,Cardoso2015A,Cardoso2017Weightless,dhamija2018reducing,yoshihashi2019classification,Shu2018Unseen,oza2019c2ae}    \\ \hline
\multirow{2}{*}{Generative model}     & Instance Generation-based     & \cite{Ge2017Generative,Neal2018Open,Jo2018Open,Yu2017Open,yang2019open}  \\ \cline{2-3}
                                      & Non-Instance Generation-based & \cite{Geng2018Collective}    \\ \hline
\end{tabular}
\end{center}
\end{table*}

Further, the authors in \cite{scheirer2012toward} also formally introduced the concept of \emph{openness} for a particular problem or data universe. 
\newtheorem{myDef}{Definition}
\begin{myDef}
\emph{(The openness defined in \cite{scheirer2012toward})} Let \emph{$C_{\text{TA}}$, $C_{\text{TR}}$, and $C_{\text{TE}}$} respectively represent the set of classes to be recognized, the set of classes used in training and the set of classes used during testing. Then the openness of the corresponding recognition task $O$ is:
\emph{
\begin{equation}
O = 1 - \sqrt{\frac{2\times |C_{\text{TR}|}}{|C_{\text{TA}}| + |C_{\text{TE}}|}},
\end{equation}}
where $|\cdot|$ denotes the number of classes in the corresponding set.
\end{myDef}
Larger \emph{openness} corresponds to more open problems, while the problem is completely closed when the \emph{openness} equals $0$.
Note that \cite{scheirer2012toward} does not explicitly give the relationships among $C_{\text{TA}}$, $C_{\text{TR}}$, and $C_{\text{TE}}$. In most existing works \cite{Scheirer2014Probability,Zhang2017Sparse,yang2019open,Geng2018Collective}, the relationship, $C_{\text{TA}} = C_{\text{TR}}\subseteq C_{\text{TE}}$, holds by default. Besides, the authors in \cite{Cardoso2017Weightless} specifically give the following relationship: $C_{\text{TA}}\subseteq C_{\text{TR}}\subseteq C_{\text{TE}}$, which contains the former case. However, such a relationship is problematic for Definition 1. Consider the following simple case: $C_{\text{TA}}\subseteq C_{\text{TR}}\subseteq C_{\text{TE}}$, and $|C_{\text{TA}}|=3, |C_{\text{TR}}|=10, |C_{\text{TE}}|=15$. Then we will have $O<0$, which is obviously unreasonable. In fact, $C_{\text{TA}}$ should be a subset of $C_{\text{TR}}$, otherwise it would make no sense because one usually does not use the classifiers trained on $C_{\text{TR}}$ to identify other classes which are not in $C_{\text{TR}}$. Intuitively, the \emph{openness} of a particular problem should only depend on the KKCs' knowledge from $C_{\text{TR}}$ and the UUCs' knowledge from $C_{\text{TE}}$ rather than $C_{\text{TA}}$, $C_{\text{TR}}$, and $C_{\text{TE}}$ their three. Therefore, in this paper, we recalibrate the formula of \emph{openness}:
\begin{equation}
O^\ast = 1 - \sqrt{\frac{2\times|C_{\text{TR}}|}{|C_{\text{TR}}| + |C_{\text{TE}}|}}.
\end{equation}

Compared to Eq. (2), Eq. (3) is just a relatively more reasonable form to estimate the \emph{openness}. Other definitions can also capture this notion, and some may be more precise, thus worth further exploring. With the concepts of \emph{open space risk} and \emph{openness} in mind, the definition of OSR problem can be given as follows:
\begin{myDef}
\emph{(The Open Set Recognition Problem\cite{scheirer2012toward})} Let $V$ be the training data, and let $R_\mathcal{O}$, $R_\varepsilon$ respectively denote the open space risk and the empirical risk. Then the goal of open set recognition is to find a measurable recognition function $f\in \mathcal{H}$, where $f(x)>0$ implies correct recognition, and $f$ is defined by minimizing the following \emph{\textbf{Open Set Risk}}:
\begin{equation}
\arg\min_{f\in \mathcal{H}}\left\{R_\mathcal{O}(f) + \lambda_rR_{\varepsilon}(f(V))\right\}
\end{equation}
where $\lambda_r$ is a regularization constant.
\end{myDef}
The open set risk denoted in formula (4) balances the empirical risk and the open space risk over the space of allowable recognition functions. Although this initial definition mentioned above is more theoretical, it provides an important guidance for subsequent OSR modeling, leading to a series of OSR algorithms which will be detailed in the following section.


\section{A Categorization of OSR Techniques}
Although Scheirer et al. \cite{scheirer2012toward} formalized the OSR problem, an important question is how to incorporate Eq. (1) to modeling. There is an ongoing debate between the use of generative and discriminative models in statistical learning \cite{Bouchard2004The,Lasserre2006Principled}, with arguments for the value of each. However, as discussed in \cite{Scheirer2014Probability}, open set recognition introduces such a new issue, in which neither discriminative nor generative models can directly address UUCs existing in open space unless some constraints are imposed. Thus, with some constraints, researchers have made the exploration in modeling of OSR respectively from the discriminative and generative perspectives. Next, we mainly review the existing OSR models from these two perspectives.

According to the modeling forms, these models can be further categorized into four categories (Table 2): Traditional ML (TML)-based and Deep Neural Network (DNN)-based methods from the discriminative model perspective; Instance and Non-Instance Generation-based methods from the generative model perspective. For each category, we review different approaches by focusing on their corresponding representative works. Moreover, a global picture on how these approaches are linked is given in Fig. 3, while several available software packages' links are also listed (Table 3) to facilitate subsequent research by relevant researchers. Next, we first give a review from the discriminative model perspective, where most existing OSR algorithms are modeled from this perspective.
\begin{figure*}[!t]
  \centering
  \includegraphics[width=15cm,height=9cm]{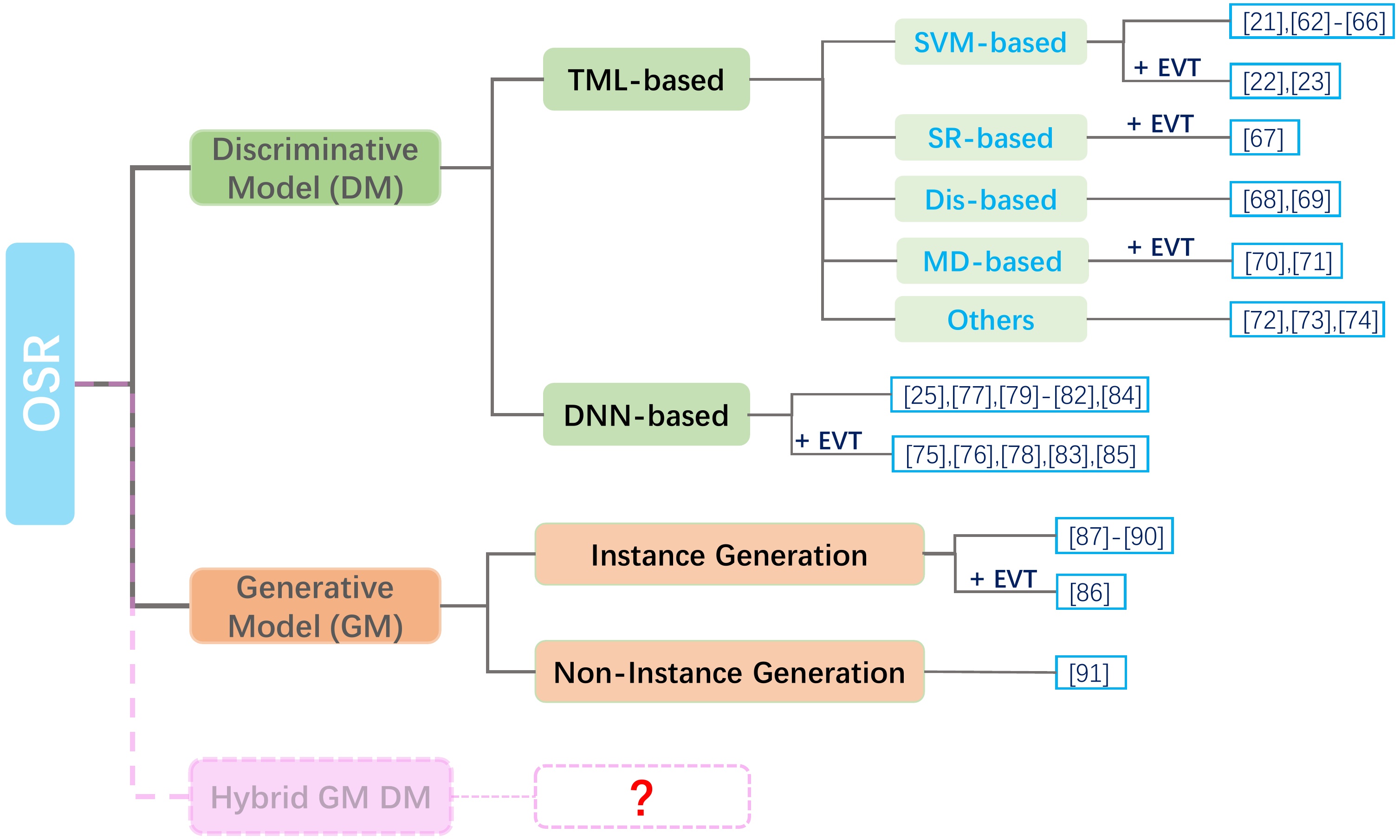}
  \caption{A global picture on how the existing OSR methods are linked. 'SR-based', 'Dis-based', 'MD-based' respectively denote the Sparse Representation-based, Distance-based and Margin Distribution-based OSR methods, while '+EVT' represents the corresponding method additionally adopts the statistical Extreme Value Theory. Note that the pink dotted line module indicates that there is no related OSR work from the hybrid generative and discriminative model perspective at present, which seems also a promising research direction in the future.}
  \label{fig_sim}
\end{figure*}

\begin{table*}[]
\tabcolsep 0mm
\renewcommand\arraystretch{1.5}
\caption{Available Software Packages}
\begin{center}
\begin{tabular}{llll}
\hline
\hline
Model \ \ & Language \ \ \ & Author & Link\\
\hline
1-vs-Set\cite{scheirer2012toward},W-SVM\cite{Scheirer2014Probability},$P_{I}$-SVM\cite{Jain2014Multi}       & C/C++ & Jain et al. & \url{https://github.com/ljain2/libsvm-openset} \\
BFHC \cite{Cevikalp2017Best}, EPCC \cite{Cevikalp2017Fast}     & Matlab & Cevikalp et al. & \url{http://mlcv.ogu.edu.tr/softwares.html}\\
SROSR \cite{Zhang2017Sparse} & Matlab & Zhang et al. & \url{https://github.com/hezhangsprinter/SROSR}\\
NNO \cite{Bendale2015Towards} & Matlab & Bendale et al. & \url{https://github.com/abhijitbendale/OWR}\\
HPLS, HFCN \cite{Vareto2018Towards}    & Python & Vareto et al. &  \url{https://github.com/rafaelvareto/HPLS-HFCN-openset}\\
\hline
OpenMax\cite{Bendale2015Towards1}     & Python & Bendale et al. & \url{https://github.com/abhijitbendale/OSDN}\\
DOC\cite{Shu2017DOC}& Python & Shu et al. & \url{https://github.com/alexander-rakhlin/CNN-for-Sentence-Classification-in-Keras}\\
RNA\cite{dhamija2018reducing}& Python & Dhamija et al. & \url{http://github.com/Vastlab/Reducing-Network-Agnostophobia}\\
\hline
OSRCI\cite{Neal2018Open}       & Python & Neal et al. & \url{https://github.com/lwneal/counterfactual-open-set}\\
ASG \cite{Yu2017Open}   & Python & Liu et al. & \url{https://github.com/eyounx/ASG}\\
\hline
EVM\cite{Rudd2018The}& Python &  Rudd et al.& \url{https://github.com/EMRResearch/ExtremeValueMachine}\\
\hline
\end{tabular}
\end{center}
\end{table*}
\subsection{Discriminative Model for OSR}
\subsubsection{Traditional ML Methods-based OSR Models}
As mentioned above, traditional machine learning methods (e.g., SVM, sparse representation, Nearest Neighbor, etc.) usually assume that the training and testing data are drawn from the same distribution. However, such an assumption does not hold any more in OSR. To adapt these methods to the OSR scenario, many efforts have been made \cite{scheirer2012toward,Cevikalp2013Face,Cevikalp2017Best,Scheirer2014Probability,Jain2014Multi,Scherreik2016Open,Cevikalp2017Polyhedral,
Cevikalp2017Fast,Zhang2017Sparse,Bendale2015Towards,J2017Nearest,Rudd2018The,Vignotto2018Extreme,Fei2016Breaking,Vareto2018Towards,Neira2018Data}.

\textbf{SVM-based}:
The Support Vector Machine (SVM) \cite{Cortes1995Support} has been successfully used in traditional classification/recognition task. However, when UUCs appear during testing, its classification performance will decrease significantly since it usually divides over-occupied space for KKCs under the closed set assumption. As shown in Fig. 2(b), once the UUCs' samples fall into the space divided for some KKCs, these samples will never be correctly classified. To overcome this problem, many SVM-based OSR methods have been proposed.

Using Definition 2, Scheirer et al. \cite{scheirer2012toward} proposed 1-vs-Set machine, which incorporates an open space risk term in modeling to account for the space beyond the reasonable support of KKCs. Concretely, they added another hyperplane parallelling the separating hyperplane obtained by SVM in score space, thus leading to a slab in feature space. The open space risk for a linear kernel slab model is defined as follows:
\begin{equation}
R_{\mathcal{O}} = \frac{\delta_{\Omega} - \delta_A}{\delta^{+}} + \frac{\delta^{+}}{\delta_{\Omega} - \delta_A} + p_A\omega_A + p_{\Omega}\omega_{\Omega},
\end{equation}
where $\delta_A$ and $\delta_{\Omega}$ denote the marginal distances of the corresponding hyperplanes, and $\delta^{+}$ is the separation needed to account for all positive data. Moreover, user-specified parameters $p_A$ and $p_{\Omega}$ are given to weight the importance between the margin spaces $\omega_A$ and $\omega_{\Omega}$. In this case, a testing sample that appears between the two hyperplanes would be labeled as the appropriate class. Otherwise, it would be considered as non-target class or rejected, depending on which side of the slab it resides. Similar to 1-vs-Set machine, Cevikalp \cite{Cevikalp2013Face,Cevikalp2017Best} added another constraint on the samples of positive/target class based on the traditional SVM, and proposed the Best Fitting Hyperplane Classifier (BFHC) model which directly formed a slab in feature space. In addition, BFHC can be extended to nonlinear case by using kernel trick, and we refer reader to \cite{Cevikalp2017Best} for more details.

Although the slab models mentioned above decrease the KKC's region for each binary SVM, the space occupied by each KKC remains unbounded. Thus the open space risk still exists. To overcome this challenge, researchers further seek new ways to control this risk \cite{Scheirer2014Probability,Jain2014Multi,Scherreik2016Open,Cevikalp2017Polyhedral,Cevikalp2017Fast}.

Scheirer et al. \cite{Scheirer2014Probability} incorporated non-linear kernels into a solution that further limited open space risk by positively labeling only sets with finite measure. They formulated a compact abating probability (CAP) model, where probability of class membership abates as points move from known data to open space. Specifically, a Weibull-calibrated SVM (W-SVM) model was proposed, which combined the statistical extreme value theory (EVT)\cite{kotz2000extreme}\footnote{Extreme Value Theory (EVT)\cite{kotz2000extreme}, also known as Fisher-Tippett Theorem, is a branch of statistics analyzing the distribution of data of abnormally high or low values. The application of EVT in visual tasks mainly involves post-recognition score analysis. For open set recognition, the threshold to reject/accept usually lies in the overlap region of extremes of match and non-match score distributions\cite{Zhang2017Sparse,oza2019c2ae}. As EVT can effectively model the tail of the match and non-match recognition scores as one of the extreme value distributions. This makes it widely used in existing OSR models. The following is the definition of EVT:\\
{\bf Extreme Value Theory}. \emph{Let $(v_1,v_2,...)$ be a sequence of i.i.d samples. Let $\zeta_n=\max\{v_1,...,v_n\}$. If a sequence of pairs of real numbers $(a_n,b_n)$ exists such that each $a_n>0$ and $\lim_{z\rightarrow\infty}P(\frac{\zeta_n-b_n}{a_n}\leq z)=F(z)$ then if $F$ is a non-degenerate distribution function, it belongs to the Gumbel, the Fr$\acute{e}$chet or the Reversed Weibull family.}} for score calibration with two separated SVMs. The first SVM is a one-class SVM CAP model used as a conditioner: if the posterior estimate $P_O(y|x)$ of an input sample $x$ predicted by one-class SVM is less than a threshold $\delta_\tau$, the sample will be rejected outright. Otherwise, it will be passed to the second SVM. The second one is a binary SVM CAP model via a fitted Weibull cumulative distribution function, yielding the posterior estimate $P_{\eta}(y|x)$ for the corresponding positive KKC. Furthermore, it also obtains the posterior estimate $P_{\psi}(y|x)$ for the corresponding negative KKCs by a reverse Weibull fitting. Defined an indicator variable: $\iota_y=1$ if $P_O(y|x)>\delta_\tau$ and $\iota_y=0$ otherwise, the W-SVM recognition for all KKCs $\mathcal{Y}$ is:
\begin{eqnarray}
\begin{split}
y^* &= \ \arg\max\limits_{y\in\mathcal{Y}} P_\eta(y|x)\times P_{\psi}(y|x)\times \iota_y  \\
    &\text{subject to}\ \  P_\eta(y^*|x)\times P_{\psi}(y^*|x) \geq \delta_R,
\end{split}
\end{eqnarray}
where $\delta_R$ is the threshold of the second SVM CAP model. The thresholds $\delta_\tau$ and $\delta_R$ are set empirically, e.g., $\delta_\tau$ is fixed to 0.001 as specified by the authors, while $\delta_R$ is recommended to set according to the openness of the specific problem
\begin{equation}
\delta_R = 0.5\times \text{openness}.
\end{equation}

Besides, W-SVM was further used for open set intrusion recognition on the KDDCUP'99 dataset \cite{Cruz2017Open}. More works on intrusion detection in open set scenario can be found in \cite{Rudd2017A}. Intuitively, we can reject a large set of UUCs (even under an assumption of incomplete class knowledge) if the positive data for any KKCs is accurately modeled without overfitting. Based on this intuition, Jain et al. \cite{Jain2014Multi} invoked EVT to model the positive training samples at the decision boundary and proposed the $P_I$-SVM algorithm. $P_I$-SVM also adopts the threshold-based classification scheme, in which the selection of corresponding threshold takes the same strategy in W-SVM.

Note that while both W-SVM and $P_I$-SVM effectively limit the open space risk by the threshold-based classification schemes, their thresholds' selection also gives some caveats. First, they are assumed that all KKCs  have equal thresholds, which may be not reasonable since the distributions of classes in feature space are usually unknown. Second, the reject thresholds are recommended to set according to the problem openness \cite{Scheirer2014Probability}. However, the openness of the corresponding problem is usually unknown as well.

To address these caveats, Scherreik et al. \cite{Scherreik2016Open} introduced the probabilistic open set SVM (POS-SVM) classifier which could empirically determine unique reject threshold for each KKC under Definition 2. Instead of defining $R_\mathcal{O}$ as relative measure of open and class-defined space, POS-SVM chooses probabilistic representations respectively for open space risk $R_{\mathcal{O}}$ and empirical risk $R_{\varepsilon}$ (details c.f. \cite{Scherreik2016Open}). Moreover, the authors also adopted a new OSR evalution metric called Youden's index which combines the true negative rate and recall, and will be detailed in subsection 5.2. Recently, to address sliding window visual object detection and open set recognition tasks, Cevikalp and Triggs \cite{Cevikalp2017Polyhedral,Cevikalp2017Fast} used a family of quasi-linear ``polyhedral conic'' functions of \cite{Rafail2006Separation} to define the acceptance regions for positive KKCs. This choice provides a convenient family of compact and convex region shapes for discriminating relatively well localized positive KKCs from broader negative ones including negative KKCs and UUCs.

\textbf{Sparse Representation-based}:
In recent years, the sparse representation-based techniques have been widely used in computer vision and image processing fields \cite{Wright2010Sparse,Rubinstein2010Dictionaries,peng2018maximum}. In particular, sparse representation-based classifier (SRC) \cite{Wright2009Robust} has gained a lot of attentions, which identifies the correct class by seeking the sparsest representation of the testing sample in terms of the training. SRC and its variants are essentially still under the closed set assumption, so in order to adapt SRC to an open environment, Zhang and Patel \cite{Zhang2017Sparse} presented the sparse representation-based open set recognition model, briefly called SROSR.

SROSR models the tails of the matched and sum of non-matched reconstruction error distributions using EVT due to the fact that most of the discriminative information for OSR is hidden in the tail part of those two error distributions. This model consists of two main stages. One stage reduces the OSR problem into hypothesis testing problems by modeling the tails of error distributions using EVT, and the other first calculates the reconstruction errors for a testing sample, then fusing the confidence scores based on the two tail distributions to determine its identity.

%

As reported in \cite{Zhang2017Sparse}, although SROSR outperformed many competitive OSR algorithms, it also contains some limitations. For example, in the face recognition task, the SROSR would fail in such cases that the dataset contained extreme variations in pose, illumination or resolution, where the self expressiveness property required by the SRC do no longer hold. Besides, for good recognition performance, the training set is required to be extensive enough to span the conditions that might occur in testing set. Note that while only SROSR is currently proposed based on sparse representation, it is still an interesting topic for future work to develop the sparse representation-based OSR algorithms.

\textbf{Distance-based}:
Similar to other traditional ML methods mentioned above, the distance-based classifiers are usually no longer valid under the open set scenario. To meet this challenge, Bendale and Boult \cite{Bendale2015Towards} established a Nearest Non-Outlier (NNO) algorithm for open set recognition by extending upon the Nearest Class Mean (NCM) classifier \cite{Mensink2013Distance,Ristin2014Incremental}. NNO carries out classification based on the distance between the testing sample and the means of KKCs, where it rejects an input sample when all classifiers reject it. What needs to be emphasized is that this algorithm can dynamically add new classes based on manually labeled data. In addition, the authors introduced the concept of open world recognition, which details in Section 4.

Further, based on the traditional Nearest Neighbor classifier, J$\acute{\text{u}}$nior et al. \cite{J2017Nearest} introduced an open set version of Nearest Neighbor classifier (OSNN) to deal with the OSR problem. Different from those works which directly use a threshold on the similarity score for the most similar class, OSNN applies a threshold on the ratio of similarity scores to the two most similar classes instead, which is called Nearest Neighbor Distance Ratio (NNDR) technique. Specifically, it first finds the nearest neighbor $t$ and $u$ of the testing sample $s$, where $t$ and $u$ come from different classes, then calculates the ratio
\begin{equation}
\text{Ratio} = d(s,t)/d(s,u),
\end{equation}
where $d(x,x')$ denotes the Euclidean distance between sample $x$ and $x'$ in feature space. If the \emph{Ratio} is less than or equal to the pre-set threshold, $s$ will be classified as the same label of $t$. Otherwise, it is considered as the UUC.

OSNN is inherently multiclass, meaning that its efficiency will not be affected as the number of available classes for training increases. Moreover, the NNDR technique can also be applied effortlessly to other classifiers based on the similarity score, e.g., the Optimum-Path Forest (OPF) classifier \cite{Papa2009Supervised}. Other metrics could be used to replace Euclidean metric as well, and even the feature space considered could be a transformed one, as suggested by the authors. Note that one limitation of OSNN is that just selecting two reference samples coming from different classes for comparison makes OSNN vulnerable to outliers \cite{Geng2018Collective}.

\textbf{Margin Distribution-based}: Considering that most existing OSR methods take little to no distribution information of the data into account and lack a strong theoretical foundation, Rudd et al. \cite{Rudd2018The} formulated a theoretically sound classifier---the Extreme Value Machine (EVM) which stems from the concept of \emph{margin distributions}. Various definitions and uses of margin distributions have been explored \cite{garg2003margin,reyzin2006boosting,aiolli2008kernel,pelckmans2008risk}, involving techniques such as maximizing the mean or median margin, taking a weighted combination margin, or optimizing the margin mean and variance. Utilizing the marginal distribution itself can provide better error bounds than those offered by a soft-margin SVM, which translates into reduced experimental error in some cases.

As an extension of margin distribution theory from a per-class formulation \cite{garg2003margin,reyzin2006boosting,aiolli2008kernel,pelckmans2008risk} to a sample-wise formulation, EVM is modeled in terms of the distribution of sample half-distances relative to a reference point. Specifically, it obtains the following theorem:
\newtheorem{thm}{\bf Theorem}
\begin{thm}\label{thm1}
Assume we are given a positive sample $x_i$ and sufficiently many negative samples $x_j$ drawn from well-defined class distributions, yielding pairwise margin estimates $m_{ij}$. Assume a continuous non-degenerate margin distribution exists. Then the distribution for the minimal values of the margin distance for $x_i$ is given by a Weibull distribution.
\end{thm}

As Theorem 1 holds for any point $x_i$, each point can estimate its own distribution of distance to the margin, thus yielding:
\newtheorem{corollary}{\bf Corollary}
\begin{corollary}(\textbf{$\Psi$ Density Function})
Given the conditions for the Theorem 1, the probability that $x'$ is included in the boundary estimated by $x_i$ is given by
\begin{equation}
\Psi(x_i,x',\kappa_i,\lambda_i) = \exp^{-\left(\frac{\|x_i-x'\|}{\lambda_i}\right)^{\kappa_i}},
\end{equation}
where $\|x_i-x'\|$ is the distance of $x'$ from sample $x_i$, and $\kappa_i,\lambda_i$ are Weibull shape and scale parameters respectively obtained from fitting to the smallest $m_{ij}$.
\end{corollary}

\textbf{Prediction}: Once EVM is trained, the probability of a new sample $x'$ associated with class $\mathcal{C}_l$, i.e., $\hat{P}(\mathcal{C}_l|x')$, can be obtained by Eq. (9), thus resulting in the following decision function
\begin{equation}
y^*=
\begin{cases}
\arg\max_{l\in\{1,...,M\}}\hat{P}(\mathcal{C}_l|x')& \text{if} \hat{P}(\mathcal{C}_l|x') \geq \delta\\
\text{"unknown"}& \text{Otherwise,}
\end{cases}
\end{equation}
where $M$ denotes the number of KKCs in training, and $\delta$ represents the probability threshold which defines the boundary between the KKCs and unsupported open space.

Derived from the margin distribution and extreme value theories, EVM has a well-grounded interpretation and can perform nonlinear kernel-free variable bandwidth incremental learning, which is further utilized to explore the open set face recognition \cite{G2017Toward} and the intrusion detection \cite{henrydoss2017incremental}. Note that it also has some limitations as reported in \cite{Vignotto2018Extreme}, in which an obvious one is that the use of geometry of KKCs is risky when the geometries of KKCs and UUCs differ. To address these limitations, Vignotto and Engelke \cite{Vignotto2018Extreme} further presented the GPD and GEV classifiers relying on approximations from EVT.


\textbf{Other Traditional ML Methods-based}:
Using center-based similarity (CBS) space learning, Fei and Liu \cite{Fei2016Breaking} proposed a novel solution for text classification under OSR scenario, while Vareto et al. \cite{Vareto2018Towards} explored the open set face recognition and proposed HPLS and HFCN algorithms by combining hashing functions, partial least squares (PLS) and fully connected networks (FCN). Neira et al. \cite{Neira2018Data} adopted the integrated idea, where different classifiers and features are combined to solve the OSR problem.  We refer the reader to \cite{Fei2016Breaking,Vareto2018Towards,Neira2018Data} for more details. As most traditional machine learning methods for classification currently are under closed set assumption, it is appealing to adapt them to the open and non-stationary environment.

\subsubsection{Deep Neural Network-based OSR Models}
Thanks to the powerful learning representation ability, Deep Neural Networks (DNNs) have gained significant benefits for various tasks such as visual recognition, Natural language processing, text classification, etc. DNNs usually follow a typical SoftMax cross-entropy classification loss, which inevitably incurs the normalization problem, making them inherently have the closed set nature. As a consequence, DNNs often make wrong predictions, and even do so too confidently, when processing the UUCs' samples. The works in \cite{Nguyen2014Deep,Goodfellow2015Explaining} have indicated that DNNs easily suffer from vulnerability to 'fooling' and 'rubbish' images which are visually far from the desired class but produce high confidence scores. To address these problems, researchers have looked at different approaches \cite{Bendale2015Towards1,Rozsa2017Adversarial,Hassen2018Learning,prakhya2017open,Shu2017DOC,kardan2017mitigating,Cardoso2015A,Cardoso2017Weightless,dhamija2018reducing,yoshihashi2019classification,Shu2018Unseen,oza2019c2ae}.

Replacing the SoftMax layer in DNNs with an OpenMax layer,  Bendale and Boult \cite{Bendale2015Towards1} proposed the OpenMax model as a first solution towards open set Deep Networks. Specifically, a deep neural network is first trained with the normal SoftMax layer by minimizing the cross entropy loss. Adopting the concept of Nearest Class Mean \cite{Mensink2013Distance,Ristin2014Incremental}, each class is then represented as a mean activation vector (MAV) with the mean of the activation vectors (only for the correctly classified training samples) in the penultimate layer of that network. Next, the training samples' distances from their corresponding class MAVs are calculated and used to fit the separate Weibull distribution for each class. Further, the activation vector's values are redistributed according to the Weibull distribution fitting score, and then used to compute a pseudo-activation for UUCs. Finally, the class probabilities of KKCs and (pseudo) UUCs are computed by using SoftMax again on these new redistributed activation vectors.

As discussed in \cite{Bendale2015Towards1}, OpenMax effectively addressed the recognition challenge for fooling/rubbish and unrelated open set images, but it fails to recognize the adversarial images which are visually indistinguishable from training samples but are designed to make deep networks produce high confidence but incorrect answers \cite{Goodfellow2015Explaining,Szegedy2014Intriguing}. Rozsa et al. \cite{Rozsa2017Adversarial} also analyzed and compared the adversarial robustness of DNNs using SoftMax layer with OpenMax: although OpenMax provides less vulnerable systems than SoftMax to traditional attacks, it is equally susceptible to more sophisticated adversarial generation techniques directly working on deep representations. Therefore, the adversarial samples are still a serious challenge for open set recognition. Furthermore, using the distance from MAV, the cross entropy loss function in OpenMax does not directly incentivize projecting class samples around the MAV. In addition to that, the distance function used in testing is not used in training, possibly resulting in inaccurate measurement in that space \cite{Hassen2018Learning}. To address this limitation, Hassen and Chan \cite{Hassen2018Learning} learned a neural network based representation for open set recognition. In this representation, samples from the same class are closed to each other while the ones from different classes are further apart, leading to larger space among KKCs for UUCs' samples to occupy.

Besides, Prakhya et al. \cite{prakhya2017open} continued to follow the technical line of OpenMax to explore the open set text classification, while Shu et al. \cite{Shu2017DOC} replaced the SoftMax layer with a 1-vs-rest final layer of sigmoids and presented Deep Open classifier (DOC) model. Kardan and Stanley \cite{kardan2017mitigating} proposed the competitive overcomplete output layer (COOL) neural network to circumvent the overgeneralization of neural networks over regions far from the training data. Based on an elaborate distance-like computation provided by a weightless neural network,  Cardoso et al. \cite{Cardoso2015A} proposed the tWiSARD algorithm for open set recognition, which is further developed in \cite{Cardoso2017Weightless}. Recently, considering the available background classes (KUCs), Dhamija et al.\cite{dhamija2018reducing} combined SoftMax with the novel Entropic Open-Set and Objectosphere losses to address the OSR problem. Yoshihashi et al. \cite{yoshihashi2019classification} presented the Classification-Reconstruction learning algorithm for open set recognition (CROSR), which utilizes latent representations for reconstruction and enables robust UUCs' detection without harming the KKCs' classification accuracy. Using class conditioned auto-encoders with novel training and testing methodology, Oza and Patel \cite{oza2019c2ae} proposed C2AE model for OSR. Compared to the works described above, Shu et al. \cite{Shu2018Unseen} paid more attention to discovering the hidden UUCs in the reject samples. Correspondingly, they proposed a joint open classification model with a sub-model for classifying whether a pair of examples belong to the same class or not, where the sub-model can serve as a distance function for clustering to discover the hidden classes in the reject samples.

\textbf{Remark:} From the discriminative model perspective, almost all existing OSR approaches adopt the threshold-based classification scheme, where recognizers in decision either reject or categorize the input samples to some KKC using empirically-set threshold. Thus the threshold plays a key role. However, at the moment, the selection for it usually depends on the knowledge from KKCs, which inevitably incurs risks due to lacking available information from UUCs\cite{Geng2018Collective}. In fact, as the KUCs' data is often available at hand \cite{da2014learning,dhamija2018reducing,masana2018metric}, we can fully leverage them to reduce such a risk and further improve the robustness of these methods for UUCs. Besides, effectively modeling the tails of the data distribution makes EVT widely used in existing OSR methods. However, regrettably, it provides no principled means of selecting the size of tail for fitting. Further, as the object frequencies in visual categories ordinarily follow long-tailed distribution \cite{reed2001pareto,zhu2014capturing}, such a distribution fitting will face challenges once the rare classes in KKCs and UUCs appear together in testing \cite{liu2019large}.
%
%
%

\subsection{Generative Model for Open Set Recognition}
In this section, we will review the OSR methods from the generative model perspective, where these methods can be further categorized into Instance Generation-based and Non-Instance Generation-based methods according to their modeling forms.

\subsubsection{Instance Generation-based OSR Models}
The adversarial learning (AL) \cite{goodfellow2016nips} as a novel technology has gained the striking successes, which employs a generative model and a discriminative model, where the generative model learns to generate samples that can fool the discriminative model as non-generated samples. Due to the properties of AL, some researchers also attempt to account for open space with the UUCs generated by the AL technique\cite{Ge2017Generative,Neal2018Open,Jo2018Open,Yu2017Open,yang2019open}.

Using a conditional generative adversarial network (GAN) to synthesize mixtures of UUCs, Ge et al. \cite{Ge2017Generative} proposed the Generative OpenMax (G-OpenMax) algorithm, which can provide explicit probability estimation over the generated UUCs, enabling the classifier to locate the decision margin according to the knowledge of both KKCs and generated UUCs. Obviously, such UUCs in their setting are only limited in a subspace of the original KKCs' space. Moreover, as reported in \cite{Ge2017Generative}, although G-OpenMax effectively detects UUCs in monochrome digit datasets, it has no significant performance improvement on natural images .

Different from G-OpenMax, Neal et al. \cite{Neal2018Open} introduced a novel dataset augmentation technique, called counterfactual image generation (OSRCI). OSRCI adopts an encoder-decoder GAN architecture to generate the synthetic open set examples which are close to KKCs, yet do not belong to any KKCs. They further reformulated the OSR problem as classification with one additional class containing those newly generated samples. Similar in spirit to \cite{Neal2018Open}, Jo et al. \cite{Jo2018Open} adopted the GAN technique to generate fake data as the UUCs' data to further enhance the robustness of the classifiers for UUCs. Yu et al. \cite{Yu2017Open} proposed the adversarial sample generation (ASG) framework for OSR. ASG can be applied to various learning models besides neural networks, while it can generate not only UUCs' data but also KKCs' data if necessary. In addition, Yang et al. \cite{yang2019open} borrowed the generator in a typical GAN networks to produce synthetic samples that are highly similar to the target samples as the automatic negative set, while the discriminator is redesigned to output multiple classes together with an UUC. Then they explored the open set human activity recognition based on micro-Doppler signatures.

\textbf{Remark}: As most Instance Generation-based OSR methods often rely on deep neural networks, they also seem to fall into the category of DNN-based methods. But please note that the essential difference between these two categories of methods lies in whether the UUCs' samples are generated or not in learning. In addition, the AL technique does not just rely on deep neural networks, such as ASG \cite{Yu2017Open}.

\subsubsection{Non-Instance Generation-based OSR Models}
Dirichlet process (DP)\cite{teh2011dirichlet,Gershman2012A,fan2013online,Teh2006Hierarchical,bargi2018adon} considered as a distribution over distributions is a stochastic process, which has been widely applied in clustering and density estimation problems as a nonparametric prior defined over the number of mixture components. This model does not overly depend on training samples and can achieve adaptive change as the data changes, making it naturally adapt to the OSR scenario.


With slight modification to hierarchical Dirichlet process (HDP), Geng and Chen \cite{Geng2018Collective} adapted HDP to OSR and proposed the collective decision-based OSR model (CD-OSR), which can address both batch and individual samples. CD-OSR first performs a co-clustering process to obtain the appropriate parameters in the training phase. In testing phase, it models each KKC's data as a group of CD-OSR using a Gaussian mixture model (GMM) with an unknown number of components/subclasses, while the whole testing set as one collective/batch is treated in the same way. Then all of the groups are co-clustered under the HDP framework. After co-clustering, one can obtain one or more subclasses representing the corresponding class. Thus, for a testing sample, it would be labeled as the appropriate KKC or UUC, depending on whether the subclass it is assigned associates with the corresponding KKC or not.



Notably, unlike the previous OSR methods, CD-OSR does not need to define the thresholds using to determine the decision boundary between KKCs and UUCs. In contrast, it introduced some threshold using to control the number of subclasses in the corresponding class, and the selection of such a threshold has been experimentally indicated more generality (details c.f. \cite{Geng2018Collective}). Furthermore, CD-OSR can provide explicit modeling for the UUCs appearing in testing, naturally resulting in a new class discovery function. Please note that such a new discovery is just at subclass level. Moreover, adopting the collective/batch decision strategy makes CD-OSR consider the correlations among the testing samples obviously ignored by other existing methods. Besides, as reported in \cite{Geng2018Collective}, CD-OSR is just as a conceptual proof for open set recognition towards collective decision at present, and there are still many limitations. For example, the recognition process of CD-OSR seems to have the flavor of lazy learning to some extent, where the co-clustering process will be repeated when other batch testing data arrives, resulting in higher computational overhead.



\textbf{Remark:} The key to Instance Generation-based OSR models is generating effective UUCs' samples. Though these existing methods have achieved some results, generating more effective UUCs' samples still need further study. Furthermore, the data adaptive property makes (hierarchical) Dirichlet process naturally suitable for dealing with the OSR task. Since only \cite{Geng2018Collective} currently gave a preliminary exploration using HDP, thus this study line is also worth further exploring. Besides, the collective decision strategy for OSR is a promising direction as well, since it not only takes the correlations among the testing samples into account but also provides a possibility for new class discovery, whereas single-sample decision strategy\footnote{The single-sample decision means that the classifier makes a decision, sample by sample. In fact, almost all existing OSR methods are designed specially for recognizing individual samples, even these samples are collectively coming in batch.} adopted by other existing OSR methods cannot do such a work since it cannot directly tell whether the single rejected sample is an outlier or from new class.

\section{Beyond Open Set Recognition}
Please note that the existing open set recognition is indeed in an open scenario but not incremental and does not scale gracefully with the number of classes. On the other hand, though new classes (UUCs) are assumed to appear incremental in class incremental learning (C-IL) \cite{cauwenberghs2001incremental,crammer2006online,fink2006online,yeh2008dynamic,muhlbaier2008learn,kuzborskij2013n}, these studies mainly focused on how to enable the system to incorporate later coming training samples from new classes instead of handling the problem of recognizing UUCs. To jointly consider the OSR and CIL tasks, Bendale and Boult \cite{Bendale2015Towards} expanded the existing open set recognition (Definition 2) to the open world recognition (OWR), where a recognition system should perform four tasks: detecting UUCs, choosing which samples to label for addition to the model, labelling those samples, and updating the classifier. Specifically, the authors give the following definition:
\begin{myDef}
\emph{(Open World Recognition\cite{Bendale2015Towards})} Let $\mathcal{K}_T\in \mathbb{N}^+$ be the set of labels of KKCs at time $T$, and let the zero label (0) be reserved for (temporarily) labeling data as unknown. Thus $\mathbb{N}$ includes the labels of KKCs and UUCs. Based on the Definition 2, a solution to open world recognition is a tuple $[F,\varphi,\nu,\mathcal{L},I]$ with:
\begin{enumerate}[\IEEEsetlabelwidth{8)}]
\item A multi-class open set recognition function $F(x): \mathbb{R}^d \mapsto \mathbb{N}$ using a vector function $\varphi(x)$ of $i$ per-class measurable recognition functions $f_i(x)$, also using a novelty detector $\nu(\varphi): \mathbb{R}_i \mapsto [0,1]$. We require the per-class recognition functions $f_i(x)\in \mathcal{H}: \mathbb{R}^d \mapsto \mathbb{R}$ for $i\in \mathcal{K}_T$ to be open set functions that manage open space risk as Eq. (1). The novelty detector $\nu(\varphi): \mathbb{R}^i \mapsto [0,1]$ determines if results from vector of recognition functions is from an UUC.
\item A labeling process $\mathcal{L}(x): \mathbb{R}^d \mapsto \mathbb{N}^{+}$ applied to novel unknown data $U_T$ from time $T$, yielding labeled data $D_T=\{(y_j,x_j)\}$ where $y_j=\mathcal{L}(x_j), \vee x_j\in U_T$. Assume the labeling finds $m$ new classes, then the set of KKCs becomes $\mathcal{K}_{T+1}=\mathcal{K}_T\cup \{i+1,...,i+m\}$.
\item An incremental learning function $I_T(\varphi;D_T): \mathcal{H}^i \mapsto \mathcal{H}^{i+m}$ to scalably learn and add new measurable functions $f_{i+1}(x)...f_{i+m}(x)$, each of which manages open space risk, to the vector $\varphi$ of measurable recognition functions.
\end{enumerate}
\end{myDef}
For more details, we refer the reader to \cite{Bendale2015Towards}. Ideally, all of these steps should be automated. However, \cite{Bendale2015Towards} only presumed supervised learning with labels obtained by human labeling at present, and proposed the NNO algorithm which has been discussed in subsection 3.1.1.

Afterward, some researchers continued to follow up this research route. Rosa et al.\cite{de2016online} argued that to properly capture the intrinsic dynamic of OWR, it is necessary to append the following aspects: (a) the incremental learning of the underlying metric, (b) the incremental estimate of confidence thresholds for UUCs, and (c) the use of local learning to precisely describe the space of classes. Towards these goals, they extended three existing metric learning methods using online metric learning. Doan and Kalita \cite{doan2017overcoming} presented the Nearest Centroid Class (NCC) model, which is similar to the online NNO\cite{de2016online} but differs with two main aspects. First, they adopted a specific solution to address the initial issue of incrementally adding new classes. Second, they optimized the nearest neighbor search for determining the nearest local balls. Lonij et al. \cite{Lonij2017Open} tackled the OWR problem from the complementary direction of assigning semantic meaning to open-world images. To handle the open-set action recognition task, Shu et al. \cite{shi2018odn} proposed the Open Deep Network (ODN) which first detects new classes by applying a multiclass triplet thresholding method, and then dynamically reconstructs the classification layer by adding predictors for new classes continually. Besides, EVM discussed in subsection 3.1.1 also adapts to the OWR scenario due to the nature of incremental learning \cite{Rudd2018The}. Recently, Xu et al. \cite{Xu2018Learning} proposed a meta-learning method to learn to accept new classes  without training under the open world recognition framework.

\textbf{Remark:} As a natural extension of OSR, OWR faces more serious challenges which require it to have not only the ability to handle the OSR task, but also minimal downtime, even to continuously learn, which seems to have the flavor of lifelong learning to some extent. Besides, although some progress regarding OWR has been made, there is still a long way to go.

\section{Datasets, Evaluation Criteria and Experiments}
\subsection{Datasets}
In open set recognition, most existing experiments are usually carried out on a variety of recast multi-class benchmark datasets at present, where some distinct labels in the corresponding dataset are randomly chosen as KKCs while the remaining ones as UUCs. Here we list some commonly used benchmark datasets and their combinations:

\textbf{LETTER}\cite{frey1991letter}: has a total of 20000 samples from 26 classes, where each class has around 769 samples with 16 features. To recast it for open set recognition, 10 distinct classes are randomly chosen as KKCs for training, while the remaining ones as UUCs.

\textbf{PENDIGITS}\cite{bilenko2004integrating}: has a total of 10992 samples from 10 classes, where each class has around 1099 samples with 16 features. Similarly, 5 distinct classes are randomly chosen as KKCs and the remaining ones as UUCs.

\textbf{COIL20}\cite{nene1996columbia}: has a total of 1440 gray images from 20 objects (72 images each object). Each image is down-sampled to $16\times16$, i.e., the feature dimension is 256. Following \cite{Geng2018Collective}, we further reduce the dimension to 55 by principal component analysis (PCA) technique, remaining 95\% of the samples' information. 10 distinct objects are randomly chosen as KKCs, while the remaining ones as UUCs.

\textbf{YALEB}\cite{georghiades2001few}: The Extended Yale B (YALEB) dataset has a total of 2414 frontal-face images from 38 individuals. Each individuals has around 64 images. The images are cropped and normalized to $32\times32$. Following \cite{Geng2018Collective}, we also reduce their feature dimension to 69 using PCA. Similar to COIL20, 10 distinct classes are randomly chosen as KKCs, while the remaining ones as UUCs.

\textbf{MNIST}\cite{lecun1998gradient}: consists of 10 digit classes, where each class contains between 6313 and 7877 monochrome images with $28\times28$ feature dimension. Following \cite{Neal2018Open}, 6 distinct classes are randomly chosen as KKCs, while the remaining 4 classes as UUCs.

\textbf{SVHN}\cite{netzer2011reading}: has ten digit classes, each containing between 9981 and 11379 color images with $32\times32$ feature dimension. Following \cite{Neal2018Open}, 6 distinct classes are randomly chosen as KKCs, while the remaining 4 classes as UUCs.

\textbf{CIFAR10}\cite{krizhevsky2009learning}: has a total of 6000 color images from 10 natural image classes. Each image has $32\times32$ feature dimension. Following \cite{Neal2018Open}, 6 distinct classes are randomly chosen as KKCs, while the remaining 4 classes as UUCs. To extend this dataset to larger openness, \cite{Neal2018Open} further proposed the \textbf{CIFAR+10}, \textbf{CIFAR+50} datasets, which use 4 non-animal classes in CIFAR10 as KKCs, while 10 and 50 animal classes are respectively chosen from CIFAR100\footnote{http://www.cs.toronto.edu/$\sim$kriz/cifar.html} as UUCs.

\textbf{Tiny-Imagenet}\cite{le2015tiny}: has a total of 200 classes with 500 images each class for training and 50 for testing, which is drawn from the Imagenet ILSVRC 2012 dataset\cite{russakovsky2015imagenet} and down-sampled to $32\times32$. Following \cite{Neal2018Open}, 20 distinct classes are randomly chosen as KKCs, while the remaining 180 classes as UUCs.

\subsection{Evaluation Criteria}
In this subsection, we summarize some commonly used evaluation metrics for open set recognition. For evaluating classifiers in the OSR scenario, a critical factor is taking the recognition of UUCs into account. Let $TP_i$, $TN_i$, $FP_i$, and $FN_i$ respectively denote the true positive, true negative, false positive, and false negative for the i-th KKC, where $i\in\{1,2,...,C\}$ and $C$ denotes the number of KKCs. Further, let $TU$ and $FU$ respectively denote the correct and false reject for UUCs. Then we can obtain the following evaluation metrics.
\subsubsection{Accuracy for OSR}
As a common choice for evaluating classifiers under closed set assumption, the accuracy $\mathcal{A}$ is usually defined as
\begin{equation*}
\mathcal{A} = \frac{\sum_{i=1}^C(TP_i + TN_i)}{\sum_{i=1}^C(TP_i + TN_i + FP_i + FN_i)}.
\end{equation*}
A trivial extension of accuracy to the OSR scenario $\mathcal{A}_{\text{O}}$ is that the correct response should contain the correct classification for KKCs and correct reject for UUCs:
\begin{equation}
\mathcal{A}_{\text{O}} = \frac{\sum_{i=1}^C(TP_i + TN_i) + TU}{\sum_{i=1}^C(TP_i + TN_i + FP_i + FN_i) + (TU + FU)}.
\end{equation}
However, as $\mathcal{A}_{\text{O}}$ denotes the sum of the correct classification for KKCs and the correct reject for UUCs, it can not objectively evaluate the OSR models. Consider the following case: when the reject performance plays the leading role, and the testing set contains large number of UUCs' samples while only a few samples for KKCs, $\mathcal{A}_{\text{O}}$ can still achieve a high value, even though the fact is that the recognizer's classification performance for KKCs is really low, and vice versa. Besides, \cite{J2017Nearest} also gave a new accuracy metric for OSR called \emph{normalized accuracy} (NA), which weights the accuracy for KKCs (AKS) and the accuracy for UUCs (AUS):
\begin{equation}
\text{NA}=\lambda_r\text{AKS} + (1-\lambda_r)\text{AUS},
\end{equation}
where
\begin{equation*}
\text{AKS} = \frac{\sum_{i=1}^C(TP_i + TN_i)}{\sum_{i=1}^C(TP_i + TN_i + FP_i + FN_i)}, \  \text{AUS} = \frac{TU}{TU+FU},
\end{equation*}
and $\lambda_r$, $0<\lambda_r<1$, is a regularization constant.

\subsubsection{F-measure for OSR}
The F-measure $F$, widely applied in information retrieval and machine learning, is defined as a harmonic mean of precision $P$ and recall $R$
\begin{equation}
F = 2 \times \frac{P\times R}{P + R}.
\end{equation}
Please note that when using F-measure for evaluating OSR classifiers, one should not consider all the UUCs appearing in testing as one additional simple class, and obtain $F$ in the same way as the multiclass closed set scenario. Because once performing such an operation, the correct classifications of UUCs' samples would be considered as true positive classifications. However, such true positive classification makes no sense, since we have no representative samples of UUCs to train the corresponding classifier. By modifying the computations of Precision and Recall only for KKCs, \cite{J2017Nearest} gave a relatively reasonable F-measure for OSR. The following equations detail these modifications, where Eq. (14) and (15) are respectively used to compute the macro-F-measure and the micro-F-measure by Eq. (13).
\begin{equation}
P_{ma}=\frac{1}{C}\sum_{i=1}^C\frac{TP_i}{TP_i+FP_i}, R_{ma}=\frac{1}{C}\sum_{i=1}^C\frac{TP_i}{TP_i+FN_i}
\end{equation}
\begin{equation}
P_{mi}=\frac{\sum_{i=1}^CTP_i}{\sum_{i=1}^C(TP_i+FP_i)}, R_{mi}=\frac{\sum_{i=1}^CTP_i}{\sum_{i=1}^C(TP_i+FN_i)}
\end{equation}

Note that although the precision and recall only consider the KKCs in Eq. (14) and (15), the $FN_i$ and $FP_i$ also consider the false UUCs and false KKCs by taking the false negative and the false positive into account (details c.f. \cite{J2017Nearest}).


\subsubsection{Youden's index for OSR}
As the F-measure is invariant to changes in $TN$ \cite{Anna2009A}, an important factor in OSR performance, Scherreik and Rigling \cite{Scherreik2016Open} turned to Youden's index $J$  defined as follows
\begin{equation}
J = R + S - 1,
\end{equation}
where $S= TN/(TN+FP)$ represents the true negative rate \cite{Youden1950Index}. Youden's index can express an algorithm's ability to avoid failure \cite{Sokolova2006Beyond}, and it is bounded in $[-1,1]$, where higher value indicates an algorithm more resistant to failure. Furthermore, the classifier is noninformative when $J=0$, whereas it tends to provide more incorrect than correct information when $J<0$.

Besides, with the aim to overcoming the effects on the sensitivity of model parameters and thresholds, \cite{Neal2018Open} adopted the area under ROC curve (AUROC) together with the closed set accuracy as the evaluation metric, which views the OSR task as a combination of novelty detection and multiclass recognition. Note that although AUROC does a good job for evaluating the models, for the OSR problem, we eventually need to make a decision (a sample belongs to which KKC or UUC), thus such thresholds seem to have to be determined.

\textbf{Remark}: Currently, F-measure and AUROC is the most commonly used evaluation metrics. As the OSR problem faces a new scenario, the new evaluation methods are worth further exploring.

\subsection{Experiments}
This subsection quantitatively assesses a number of representative OSR methods on the popular benchmark datasets mentioned in subsection 5.1. Further, these methods are compared in terms of the classification of non-depth and depth features.


\subsubsection{OSR Methods Using Non-depth Feature}
The OSR methods using non-depth feature are usually evaluated on LETTER, PENDIGITS, COIL20, YALEB datasets. Most of them adopt the threshold-based strategy, where the thresholds are recommended to set according to the openness of the specific problem \cite{Scheirer2014Probability,Jain2014Multi,Zhang2017Sparse}. However, we usually have no prior knowledge about UUCs in the OSR scenario. Thus such a setting seems unreasonable, which is recalibrated in this paper, i.e., the decision thresholds are determined only based on the KKCs in training, and once they are determined during training, their values will no longer change in testing. To effectively determine the thresholds and parameters of the corresponding models, we introduce an evaluation protocol referring to \cite{J2017Nearest,Geng2018Collective}, as follows.

\textbf{Evaluation Protocol}: As shown in Fig. 4, the dataset is first divided into training set owning KKCs and testing set containing KKCs and UUCs, respectively. 2/3 of the KKCs occurring in training set are chosen as the "KKCs" simulation, while the remaining as the "UUCs" simulation. Thus the training set is divided into a fitting set $\mathcal{F}$ just containing 'KKCs' and a validation set $\mathcal{V}$ including a 'Closed-Set' simulation and an 'Open-Set' simulation. The 'Closed-Set' simulation only owns KKCs, while the 'Open-Set' simulation contains "KKCs" and "UUCs". Note that in the training phase, all of the methods are trained with $\mathcal{F}$ and evaluated on $\mathcal{V}$. Specifically, for each experiment, we
\begin{enumerate}[\IEEEsetlabelwidth{8)}]
\item [1.] randomly select $\Omega$ distinct classes as KKCs for training from the corresponding dataset;
\item [2.] randomly choose 60\% of the samples in each KKC as training set;
\item [3.] select the remaining 40\% of the samples from step 2 and the samples from other classes excluding the $\Omega$ KKCs as testing set;
\item [4.] randomly select $[(\frac{2}{3}\Omega+0.5)]$ classes as "KKCs" for fitting from the training set, while the remaining classes as "UUCs" for validation;
\item [5.] randomly choose 60\% of the samples from each "KKC" as fitting set $\mathcal{F}$;
\item [6.] select the remaining 40\% of the samples from step 5 as the 'Closed Set' simulation, while the remaining 40\% of the samples from step 5 and the ones from "UUCs" as the 'Open-Set' simulation;
\item [7.] train the models with $\mathcal{F}$ and verify them on $\mathcal{V}$, then find the suitable model parameters and thresholds;
\item [8.] evaluate the models with 5 random class partitions using micro-F-measure.
\end{enumerate}
\textbf{Note that} the experimental protocol here is just a relatively reasonable form to evaluate the OSR methods. In fact, other protocols can also use for evaluating, and some may be more suitable, thus worth further exploring. Furthermore, since different papers often adopted different evaluation protocol before, to the best of our ability we here try to follow the parameter tuning principles in their papers. In addition, to encourage reproducible research, we refer the reader to our github\footnote{\url{https://github.com/ChuanxingGeng/Open-Set-Recognition}} for the details about the datasets and their corresponding class partitions.

\begin{figure}[!t]
  \centering
  \includegraphics[width=0.4\textwidth]{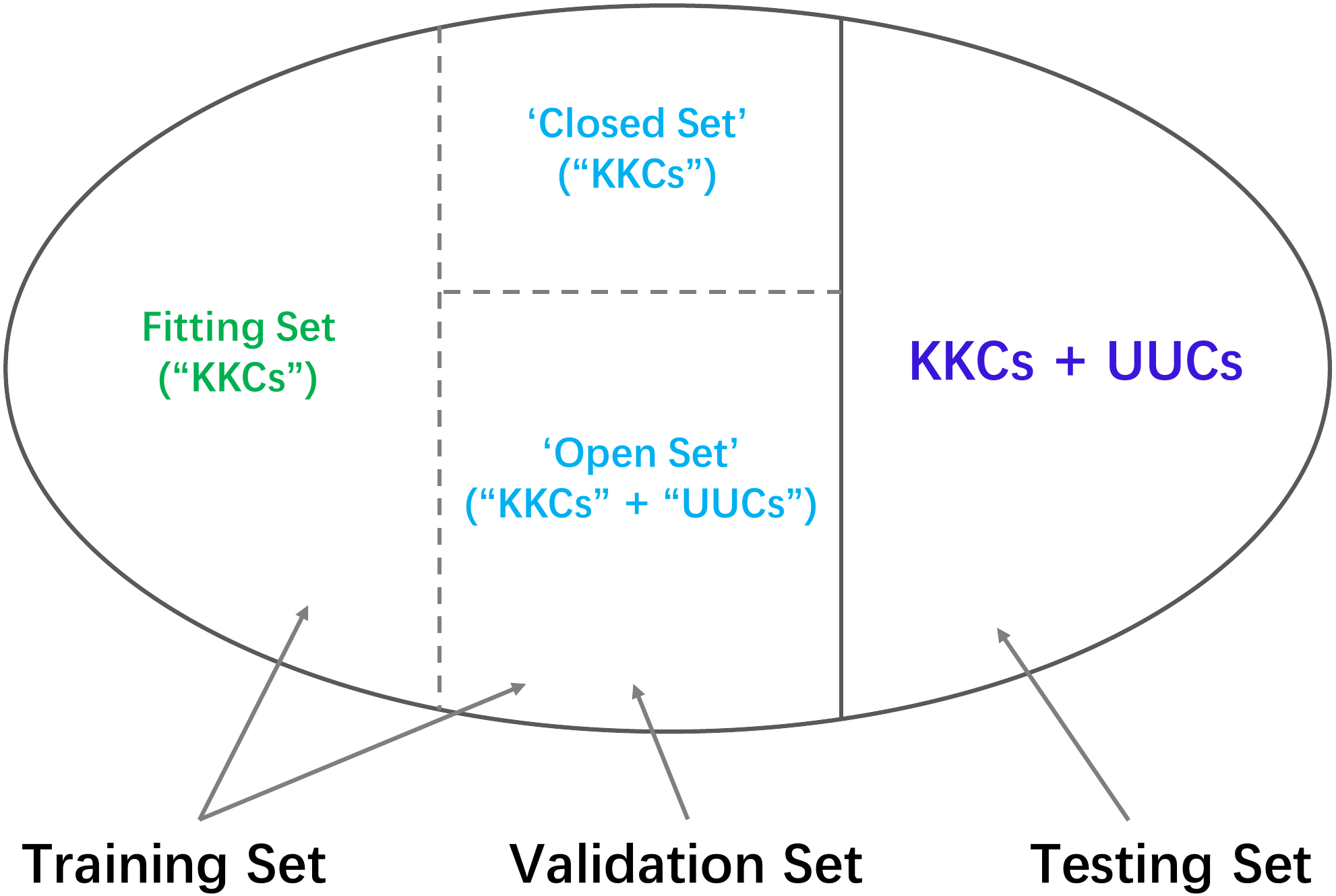}
  \caption{Data Split. The dataset is first divided into training and testing set, then the training set is further divided into a fitting set and a validation set containing a 'closed set' simulation and an 'open set' simulation.}
  \label{fig_sim}
\end{figure}

Under different openness $O^*$, Table 4 reports the comparisons among these methods, where 1-vs-Set \cite{scheirer2012toward}, W-SVM (W-OSVM\footnote{W-OSVM here denotes only using the one-clss SVM CAP model, which can be seen as benchmark of one-class classifier}) \cite{Scheirer2014Probability}, $P_I$-SVM \cite{Jain2014Multi}, SROSR \cite{Zhang2017Sparse}, OSNN \cite{J2017Nearest}, EVM \cite{Rudd2018The} from Traditional ML-based category and CD-OSR \cite{Geng2018Collective} from the Non-Instance Generation-based category.

{\bf Our first observation is:} With the increase of openness, although threshold-based methods (like W-SVM, $P_I$-SVM, SROSR, EVM) perform well on some datasets, there are also cases of significant performance degradation on other datasets (e.g., W-SVM performs well on LETTER, whereas underperformed on PENDIGITS significantly). This is mainly due to the fact that their decision thresholds are selected only based on the knowledge of KKCs, where once the UUCs' samples fall into the space divided for some KKCs, the OSR risk will be incurred. In contrast, due to the data adaptation characteristic of HDP, CD-OSR can effectively model the UUCs appearing in testing, making it currently achieve better performance on most datasets, especially for LETTER and PENDIGITS.

{\bf Our second observation is:} Compared with the other methods, the performance of OSNN fluctuates greatly in terms of the standard deviation, especially for LETTER, which is likely because the NNDR strategy makes its performance heavily dependent on the distribution characteristics of the corresponding datasets. Furthermore, as the open space in 1-vs-Set is still unbounded, we can see that its performance drops sharply with the increase of openness. As a benchmark of one-class classifier, W-OSVM here works well in the closed set scenario. However, once the scenario turns to the open set, its performance also drops significantly.

\textbf{A Summary}: Overall, based on the data adaptation characteristic of HDP, CD-OSR currently performs relatively well compared with the other methods. However, CD-OSR is also limited by HDP itself, such as difficulty in applying it to high-dimensional data, high computational complexity, etc. As for the other methods, they are limited by the underlying models they adopt as well. For example, as SRC does not work well on LETTER, thus SROSR obtains poor performance on that dataset. Furthermore, as mentioned in the remark of subsection 3.1, for the methods using EVT such as W-SVM, $P_I$-SVM, SROSR, EVM, they may face challenges once the rare classes in KKCs and UUCs appear together in testing. Additionally, it is also necessary to point out that this part just gives a comparison of these algorithms on all commonly used datasets, which may not fully characterize their behavior to some extent.

\begin{table*}[]
\centering
\caption{Comparison Among The Representative OSR Methods Using Non-depth Features}
\tabcolsep 1.6mm
\renewcommand\arraystretch{1.935}
\begin{tabular}{cl|c|c|c|c|c|c|c||c}
\hline
\multicolumn{2}{c|}{{ \textbf{Dataset / Method}}} & \textbf{1-vs-Set} & \textbf{W-OSVM} & \textbf{W-SVM} & \textbf{$P_I$-SVM} & \textbf{SROSR} & \textbf{OSNN} & \textbf{EVM} & \textbf{CD-OSR} \\ \hline
\multirow{3}{*}{\textbf{LETTER}}        &$O^*$=0\% &81.51$\pm$3.94 &95.64$\pm$0.37 &95.64$\pm$0.25 &\underline{96.92$\pm$0.36} &84.21$\pm$2.49&83.12$\pm$17.41 &96.59$\pm$0.50 &\textbf{96.94$\pm$1.36} \\
                               &$O^*$=15.48\% &55.43$\pm$3.18 &83.83$\pm$2.85 &\underline{91.24$\pm$1.48} &90.89$\pm$1.80 &74.36$\pm$5.10   &73.20$\pm$15.21 &89.81$\pm$0.40 &\textbf{91.51$\pm$1.58}        \\
                               &$O^*$=25.46\% &42.08$\pm$2.63 &73.37$\pm$1.67 &\underline{85.72$\pm$0.85} &84.16$\pm$1.01 &66.50$\pm$8.22 &64.97$\pm$13.75 &82.81$\pm$2.42 &\textbf{86.21$\pm$1.46}       \\ \hline
\multirow{3}{*}{\textbf{PENDIGITS}}     &$O^*$=0\% &97.17$\pm$0.58 &94.84$\pm$1.46 &98.82$\pm$0.26 &\textbf{99.21$\pm$0.29} &97.43$\pm$0.93   &98.55$\pm$0.71 &98.42$\pm$0.73 &\underline{99.16$\pm$0.25}        \\
                               &$O^*$=8.71\%  &78.43$\pm$1.93 &87.22$\pm$1.71 &93.05$\pm$1.85 &92.38$\pm$2.68 &96.33$\pm$1.59       &95.55$\pm$1.30&\underline{96.97$\pm$1.37} &\textbf{98.75$\pm$0.65}        \\
                               &$O^*$=18.35\% &61.29$\pm$2.52 &78.55$\pm$4.91 &88.39$\pm$3.14 &87.60$\pm$4.78 &\underline{93.53$\pm$3.26}       &90.11$\pm$4.15&92.88$\pm$2.79&\textbf{98.43$\pm$0.73}       \\ \hline
\multirow{3}{*}{\textbf{COIL20}} &$O^*$=0\% &89.59$\pm$1.81 &93.94$\pm$1.87 &86.83$\pm$1.82 &89.30$\pm$1.45 &97.12$\pm$0.60&79.61$\pm$7.41&\underline{97.68$\pm$0.88} &\textbf{97.71$\pm$0.94}        \\
                               &$O^*$=10.56\% &70.21$\pm$1.67 &90.82$\pm$2.31 &85.64$\pm$2.47 &87.68$\pm$2.02 &\underline{96.68$\pm$0.32} &73.01$\pm$6.18&95.69$\pm$1.46 &\textbf{97.32$\pm$1.50}        \\
                               &$O^*$=18.35\% &57.72$\pm$1.50 &87.97$\pm$5.40 &84.54$\pm$3.79 &86.22$\pm$3.34 &\textbf{96.45$\pm$0.66} &66.18$\pm$4.49&93.62$\pm$3.33 &\underline{95.12$\pm$2.14 }       \\ \hline
\multirow{3}{*}{\textbf{YALEB}} &$O^*$=0\% &87.99$\pm$2.42 &82.60$\pm$3.54 &86.01$\pm$2.42 &\textbf{93.47$\pm$2.74} &88.09$\pm$3.41 &81.81$\pm$8.40&68.94$\pm$6.47 &\underline{89.75$\pm$1.15}        \\
                               &$O^*$=23.30\% &49.36$\pm$1.96 &63.43$\pm$5.33 &84.56$\pm$2.19 &\textbf{88.96$\pm$1.16} &83.99$\pm$4.19 &72.90$\pm$9.41&54.40$\pm$5.77 &\underline{88.00$\pm$2.19}        \\
                               &$O^*$=35.45\% &34.37$\pm$1.44 &55.40$\pm$5.26 &83.44$\pm$2.02 &\textbf{86.63$\pm$0.60} &81.38$\pm$5.26 &67.24$\pm$7.29&46.64$\pm$5.40 &\underline{85.56$\pm$1.07 }       \\ \hline
\end{tabular}

\vspace{0.1cm}
\emph{The results report the averaged micro-F-measure (\%) over 5 random class partitions. Best and the second best performing methods are highlighted in bold \\and underline, respectively. $O^*$ calculated from Eq. (3) denotes the openness of the corresponding dataset.}
\end{table*}

\subsubsection{OSR Methods Using Depth Feature}
The OSR methods using depth feature are often evaluated on MNIST, SVHN, CIFAR10, CIFAR+10, CIFAR+50, Tiny-Imagenet. As most of them followed the evaluation protocol\footnote{The datasets and class partitions in \cite{Neal2018Open} can be find in \url{https://github.com/lwneal/counterfactual-open-set}.} defined in \cite{Neal2018Open} and did not provide source codes, similar to \cite{Pan2010A,wang2018deep}, we here only compare with their published results. Table 5 summaries the  comparisons among these methods, where SoftMax \cite{Neal2018Open}, OpenMax \cite{Bendale2015Towards1}, CROSR \cite{yoshihashi2019classification}, and C2AE \cite{oza2019c2ae} from Deep Neural Network-based category, while G-OpenMax \cite{Ge2017Generative} and OSRCI \cite{Neal2018Open}  from Instance Generation-based category.

\begin{table*}[]
\centering
\caption{Comparison Among The Representative OSR Methods Using Depth Features}
\tabcolsep 1.8mm
\renewcommand\arraystretch{1.9}
\begin{tabular}{cl|c|c|c|c||c|c}
\hline
\multicolumn{2}{c|}{\textbf{Dataset / Method }}& \textbf{SoftMax} & \textbf{OpenMax} & \textbf{CROSR} & \textbf{C2AE} & \textbf{G-OpenMax} & \textbf{OSRCI} \\ \hline
\textbf{MNIST}   &$O^*$=13.40\% &97.8 &98.1 &\textbf{99.8} &\underline{98.9} &98.4 &98.8       \\ \hline
\textbf{SVHN}    &$O^*$=13.40\% &88.6 &89.4 &\textbf{95.5} &\underline{92.2} &89.6 &91.0       \\ \hline
\textbf{CIFAR10} &$O^*$=13.40\% &67.7 &69.5 &---     &\textbf{89.5} &67.5 &\underline{69.9}       \\ \hline
\textbf{CIFAR+10}&$O^*$=24.41\% &81.6 &81.7 &---     &\textbf{95.5} &82.7 &\underline{83.8}       \\ \hline
\textbf{CIFAR+50}&$O^*$=61.51\% &80.5 &79.6 &---     &\textbf{93.7} &81.9 &\underline{82.7}       \\ \hline
\textbf{TinyImageNet}&$O^*$=57.36\% &57.7 &57.6 &\underline{67.0}   &\textbf{74.8} &58.0 &58.6       \\ \hline
\end{tabular}

\vspace{0.1cm}
\emph{The results report the averaged area under the ROC curve (\%) over 5 random class partitions\cite{Neal2018Open}. Best and the \\ second best performing methods are highlighted in bold and underline, respectively. $O^*$ calculated from Eq. (3) \\denotes the openness of the corresponding dataset. Following \cite{oza2019c2ae}, we here only copy the AUROC values of these \\methods as some of the results do not provide standard deviations.}
\end{table*}

{\bf Our first observation is:} First, the performance of all methods on MNIST are comparable, which is mainly because results on MNIST are almost saturated. Second, compared with the earlier methods like SoftMax, OpenMax, G-OpenMax and OSRCI, CROSR and C2AE currently achieve better performance on  the benchmark datasets. The main reason for their successes perhaps are: for CROSR, training networks for joint classification and reconstruction of KKCs makes the representation learned for KKCs more discriminative and tight (making the KKCs obtain tighter distribution areas); for C2AE, dividing OSR into closed set classification and open set identification allows it to avoid performing these two subtasks, concurrently, under a single score modified by the SoftMax scores (finding such a single score measure usually is extremely challenging \cite{oza2019c2ae}).

{\bf Our second observation is:} As a state-of-the-art Instance Generation-based OSR method, OSRCI currently does not win CROSR and C2AE (two state-of-the-art Deep Neural Network-based OSR methods) on almost all datasets mentioned above, this seems a bit counter-intuition, since OSRCI gains the additional information from UUCs. But this just right indicates (from another side) that the performance of Instance Generation-based methods still have more room for improvement, deserving further exploration, while also showing the effectiveness of the strategies in CROSR and C2AE.


{\bf Remark}: As mentioned previously, due to using EVT, OpenMax, CROSR, C2AE and G-OpenMax may also face challenges when the rare classes in KKCs and UUCs appear together in testing. In addition, it is also worth mentioning that the Instance Generation-based methods are orthogonal to the other three categories of methods, meaning that it can be combined with those methods to achieve their best.



\section{Future Research Directions}
In this section, we briefly analyze and discuss the limitations of the existing OSR models, while some promising research directions in this field are also pointed out and detailed in the following aspects.
\subsection{About Modeling}
First, as shown in Fig. 3, while almost all existing OSR methods are modeled from the discriminative or generative model perspective, a natural question is: can construct OSR models from the hybrid generative discriminative model perspective? Note that to our best knowledge, there is no OSR work from this perspective at the moment, which deserves further discussions. Second, the main challenge for OSR is that the traditional classifiers under closed set scenario dive over-occupied space for KKCs, thus once the UUCs' samples fall into the space divided for KKCs, they will never be correctly classified. From this viewpoint, the following two modeling perspectives will be promising research directions.


\subsubsection{Modeling known known classes}
To moderate the over-occupied space problem above, we usually expect to obtain better discrimination for each target class while confining it to a compact space with the help of clustering methods. To achieve this, the clustering and classification learning can be unified to achieve the best of both worlds: the clustering learning can help the target classes obtain tighter distribution areas (i.e., limited space), while the classification learning provides better discriminativeness for them. In fact, there have been some works fusing the clustering and classification functions into a unified learning framework \cite{Cai2010A,Qian2012Simultaneous}. Unfortunately, these works are still under a closed set assumption. Thus some serious efforts need to be done to adapt them to the OSR scenario or to specially design this type of classifiers for OSR.

\subsubsection{Modeling unknown unknown classes}
Under the open set assumption, modeling UUCs is impossible, as we only have the available knowledge from KKCs. However, properly relaxing some restrictions will make it possible, where one way is to generate the UUCs' data by the adversarial learning technique to account for open space to some extent like \cite{Neal2018Open,Jo2018Open,Yu2017Open}, in which the key is how to generate the valid UUCs' data. Besides, due to the data adaptive nature of Dirichlet process, the Dirichlet process-based OSR methods, such as CD-OSR \cite{Geng2018Collective}, are worth for further exploration as well.


\subsection{About Rejecting}
Until now, most existing OSR algorithms mainly care about effectively rejecting UUCs, yet only a few works \cite{Bendale2015Towards,Shu2018Unseen} focus on the subsequent processing for the reject samples, and these works usually adopt a post-event strategy\cite{Geng2018Collective}. Therefore, expanding existing open set recognition together with new class knowledge discovery will be an interesting research topic. Moreover, to our best knowledge, the interpretability of reject option seems to have not been discussed as well, in which a reject option may correspond to a low confidence target class, an outlier, or a new class, which is also an interesting future research direction. Some related works in other research communities can be found in \cite{Mu2017Classification,Liu2018Open,Vyas2018Out,masana2018metric,dong2018learning}.

\subsection{About the Decision}
As discussed in subsection 3.2.2, almost all existing OSR techniques are designed specially for recognizing individual samples, even
these samples are collectively coming in batch like image-set recognition \cite{wang2017joint}. In fact, such a decision does not consider correlations among the testing samples. Therefore, the collective decision \cite{Geng2018Collective} seems to be a better alternative as it can not only take the correlations among the testing samples into account but also make it possible to discover new classes at the same time. We thus expect a future direction on extending the existing OSR methods by adopting such a collective decision.

\subsection{Open Set + Other Research Areas}
As open set scenario is a more practical assumption for the real-world classification/recognition tasks, it can naturally be combined with various fields involving classification/recognition such as semi-supervised learning, domain adaptation, active learning, multi-task learning, multi-view learning, multi-label image classification problem, and so forth. For example, \cite{Busto2017Open,Saito2018Open,Baktashmotlagh2018Learning} have introduced this scenario into domain adaptation, while \cite{pham2018bayesian} introduced it to the semantic instance segmentation task. Recently, \cite{Liu2019Active} explored the open set classification in active learning field. It is also worthy mentioning that the datasets NUS-Wide and MS COCO have been used for studying multi-label zero-shot learning \cite{lee2018multi}, which are suitable to the study of multi-label OSR problem as well. Therefore, many interesting works are worth looking forward to.

\subsection{Generalized Open Set Recognition}
OSR assumes that only the KKCs' knowledge is available in training, meaning that we can also utilize a variety of side-information regarding the KKCs. Nevertheless most existing OSR methods just use the feature level information of KKCs, leaving out their other side-information such as semantic/attribute information, knowledge graph, the KUCs' data (e.g., the universum data), etc, which is also important for further improving their performances. Therefore, we give the following promising research directions.

%

\subsubsection{Appending semantic/attribute information}
With the exploration of ZSL, we can find that a lot of semantic/attibute information is usually shared between KKCs and unknown class data. Therefore, such information can fully be used to 'cognize' UUCs in OSR, or at least to provide a rough semantic/attribute description for the UUCs' samples instead of simply rejecting them. Note that this setup is different from the one in ZSL (or G-ZSL) which assumes that the semantic/attibute information of both the KKCs and UUCs are known in training. Furthermore, the last row of Table 1 shows this difference. Besides, some related works can be found in \cite{Fu2016Semi,Fu2017Vocabulary,Lonij2017Open,dong2018learning}. There also some conceptually similar topics have been studied in other research communities such as open-vocabulary object retrieval \cite{guadarrama2014open,guadarrama2016understanding}, open world person re-identification \cite{zheng2016towards} or searching targets \cite{sattar2015prediction}, open vocabulary scene parsing \cite{Zhao2017Open}.


\subsubsection{Using other available side-information}
For the over-occupied space problem mentioned in subsection 6.1, the open space risk will also be reduced as the space divided for those KKCs decreases by using other side-information like the KUCs data (e.g., universum data \cite{Cherkassky2011Practical,Qi2012Twin}) to shrink their regions as much as possible. As shown in Fig.1, taking the digital identification as an example, assume the training set including the classes of interest '1', '3', '4', '5', '9'; the testing set including all of the classes '0'-'9'. If we also have the available universum data---English letters 'Z', 'I', 'J', 'Q', 'U', we can fully use them in modeling to extend the existing OSR models, further reducing the open space risk. We therefore foresee a more generalized setting will be adopted by the future open set recognition.

\subsection{Relative Open Set Recognition}
While the open set scenario is ubiquitous, there are also some real-world scenarios that are not completely open in practice. Recognition/classification in such scenarios can be called relative open set recognition. Taking the medical diagnosis as an example, the whole sample space can be divided into two subspace respectively for sick and healthy samples, and at such a level of detecting whether the sample is sick or not, it is indeed a closed set problem. However, when we need to further identify the types of the diseases, this will naturally become a complete OSR problem since new disease unseen in training may appear in testing. There are few works currently exploring this novel mixed scenario jointly. Please note that under such a scenario, the main goal is to limit the scope of the UUCs appearing in testing, while finding the most specific class label of a novel sample on the taxonomy built with KKCs. Some related work can be found in \cite{lee2018hierarchical}.

\subsection{Knowledge Integration for Open Set Recognition}
In fact, the incomplete knowledge of the world is universal, especially for single individuals: something you know does not mean I also know. For example, the terrestrial species (sub-knowledge set) obviously are the open set for the classifiers trained on marine species. As the saying goes, "\emph{two heads are better than one}", thus how to integrate the classifiers trained on each sub-knowledge set to further reduce the open space risk will be an interesting yet challenging topic in the future work, especially for such a situation: we can only obtain the classifiers trained on corresponding sub-knowledge sets, yet these sub-knowledge sets are not available due to the data privacy protection. This seems to have the flavor of domain adaptation having multiple source domains and one target domain (mS1T) \cite{Bendavid2010A,Sun2015A,Liu2017Structure,Yu2018Unsupervised} to some extent.

\section{Conclusion}
As discussed above, in real-world recognition/classification tasks, it is usually impossible to model everything \cite{dietterich2017steps}, thus the OSR scenario is ubiquitous. On the other hand, although many related algorithms have been proposed for OSR, it still faces serious challenges. As there is no systematic summary on this topic at present, this paper gives a comprehensive review of existing OSR techniques, covering various aspects ranging from related definitions, representations of models, datasets, evaluation criteria, and algorithm comparisons. Note that for the sake of convenience, the categorization of existing OSR techniques in this paper is just one of the possible ways, while other ways can also effectively categorize them, and some may be more appropriate but beyond our focus here.


Further, in order to avoid the reader confusing the tasks similar to OSR, we also briefly analyzed the relationships between OSR and its related tasks including zero-shot, one-shot (few-shot) recognition/learning techniques, classification with reject option, and so forth. Beyond this, as a natural extension of OSR, the open world recognition was reviewed as well. More importantly, we analyzed and discussed the limitations of these existing approaches, and pointed out some promising subsequent research directions in this field.


%
%
%
%

%

\ifCLASSOPTIONcompsoc
  \section*{Acknowledgments}
\else
  \section*{Acknowledgment}
\fi

The authors would like to thank the support from the Key Program of NSFC under Grant No. 61732006, NSFC under Grant No. 61672281, and the Postgraduate Research \& Practice Innovation Program of Jiangsu Province under Grant No. KYCX18\_0306.

\ifCLASSOPTIONcaptionsoff
  \newpage
\fi



%
%
%

\bibliographystyle{ieeetr}
\bibliography{mybibfile}

\begin{thebibliography}{100}

\bibitem{Thrun1995Lifelong}
S.~Thrun and T.~M. Mitchell, ``Lifelong robot learning,'' {\em Robotics and
  Autonomous Systems}, vol.~15, no.~1-2, pp.~25--46, 1995.

\bibitem{Pentina2014A}
A.~Pentina and C.~H. Lampert, ``A pac-bayesian bound for lifelong learning,''
  in {\em International Conference on International Conference on Machine
  Learning}, pp.~II--991, 2014.

\bibitem{Pan2010A}
S.~J. Pan and Q.~Yang, ``A survey on transfer learning,'' {\em IEEE
  Transactions on Knowledge and Data Engineering}, vol.~22, no.~10,
  pp.~1345--1359, 2010.

\bibitem{Weiss2016A}
K.~Weiss, T.~M. Khoshgoftaar, and D.~D. Wang, ``A survey of transfer
  learning,'' {\em Journal of Big Data}, vol.~3, no.~1, p.~9, 2016.

\bibitem{shao2015transfer}
L.~Shao, F.~Zhu, and X.~Li, ``Transfer learning for visual categorization: A
  survey,'' {\em IEEE Transactions on Neural Networks and Learning Systems},
  vol.~26, no.~5, pp.~1019--1034, 2015.

\bibitem{Patel2015Visual}
V.~M. Patel, R.~Gopalan, R.~Li, and R.~Chellappa, ``Visual domain adaptation: A
  survey of recent advances,'' {\em IEEE Signal Processing Magazine}, vol.~32,
  no.~3, pp.~53--69, 2015.

\bibitem{yamada2014domain}
M.~Yamada, L.~Sigal, and Y.~Chang, ``Domain adaptation for structured
  regression,'' {\em International Journal of Computer Vision}, vol.~109,
  no.~1-2, pp.~126--145, 2014.

\bibitem{Palatucci2009Zero}
M.~Palatucci, D.~Pomerleau, G.~Hinton, and T.~M. Mitchell, ``Zero-shot learning
  with semantic output codes,'' in {\em International Conference on Neural
  Information Processing Systems}, pp.~1410--1418, 2009.

\bibitem{Lampert2009Learning}
C.~H. Lampert, H.~Nickisch, and S.~Harmeling, ``Learning to detect unseen
  object classes by between-class attribute transfer,'' {\em in Proceedings of
  the IEEE Conference on Computer Vision and Pattern Recognition},
  pp.~951--958, 2009.

\bibitem{fu2018recent}
Y.~Fu, T.~Xiang, Y.-G. Jiang, X.~Xue, L.~Sigal, and S.~Gong, ``Recent advances
  in zero-shot recognition: Toward data-efficient understanding of visual
  content,'' {\em IEEE Signal Processing Magazine}, vol.~35, no.~1,
  pp.~112--125, 2018.

\bibitem{Li2006One}
F.~F. Li, R.~Fergus, and P.~Perona, ``One-shot learning of object categories,''
  {\em IEEE Transactions on Pattern Analysis and Machine Intelligence},
  vol.~28, no.~4, pp.~594--611, 2006.

\bibitem{Lake2013One}
B.~M. Lake, R.~Salakhutdinov, and J.~B. Tenenbaum, ``One-shot learning by
  inverting a compositional causal process,'' in {\em International Conference
  on Neural Information Processing Systems}, pp.~2526--2534, 2013.

\bibitem{Li2008A}
F.~F. Li, Fergus, and Perona, ``A bayesian approach to unsupervised one-shot
  learning of object categories,'' in {\em IEEE International Conference on
  Computer Vision, 2003. Proceedings}, pp.~1134--1141 vol.2, 2008.

\bibitem{Amit2007Uncovering}
Y.~Amit, M.~Fink, N.~Srebro, and S.~Ullman, ``Uncovering shared structures in
  multiclass classification,'' in {\em Machine Learning, Proceedings of the
  Twenty-Fourth International Conference}, pp.~17--24, 2007.

\bibitem{Torralba2010Using}
A.~Torralba, K.~P. Murphy, and W.~T. Freeman, ``Using the forest to see the
  trees: Exploiting context for visual object detection and localization,''
  {\em Communications of the Acm}, vol.~53, no.~3, pp.~107--114, 2010.

\bibitem{fu2013learning}
Y.~Fu, T.~M. Hospedales, T.~Xiang, and S.~Gong, ``Learning multimodal latent
  attributes,'' {\em IEEE transactions on pattern analysis and machine
  intelligence}, vol.~36, no.~2, pp.~303--316, 2013.

\bibitem{vinyals2016matching}
O.~Vinyals, C.~Blundell, T.~Lillicrap, D.~Wierstra, {\em et~al.}, ``Matching
  networks for one shot learning,'' in {\em Advances in neural information
  processing systems}, pp.~3630--3638, 2016.

\bibitem{Snell2017Prototypical}
J.~Snell, K.~Swersky, and R.~S. Zemel, ``Prototypical networks for few-shot
  learning,'' {\em In Advances in Neural Information Processing Systems},
  pp.~4077--4087, 2017.

\bibitem{Bertinetto2016Learning}
L.~Bertinetto, J.~F. Henriques, J.~Valmadre, P.~H.~S. Torr, and A.~Vedaldi,
  ``Learning feed-forward one-shot learners,'' {\em In Advances in Neural
  Information Processing Systems}, pp.~523--531, 2016.

\bibitem{Chen2018Semantic}
Z.~Chen, Y.~Fu, Y.~Zhang, Y.~G. Jiang, X.~Xue, and L.~Sigal, ``Semantic feature
  augmentation in few-shot learning,'' {\em arXiv preprint arXiv:1804.05298},
  2018.

\bibitem{scheirer2012toward}
W.~J. Scheirer, A.~de~Rezende~Rocha, A.~Sapkota, and T.~E. Boult, ``Toward open
  set recognition,'' {\em IEEE transactions on pattern analysis and machine
  intelligence}, vol.~35, no.~7, pp.~1757--1772, 2013.

\bibitem{Scheirer2014Probability}
W.~J. Scheirer, L.~P. Jain, and T.~E. Boult, ``Probability models for open set
  recognition,'' {\em IEEE Transactions on Pattern Analysis and Machine
  Intelligence}, vol.~36, no.~11, pp.~2317--2324, 2014.

\bibitem{Jain2014Multi}
L.~P. Jain, W.~J. Scheirer, and T.~E. Boult, ``Multi-class open set recognition
  using probability of inclusion,'' in {\em European Conference on Computer
  Vision}, pp.~393--409, 2014.

\bibitem{Ross2010Known}
N.~A. Ross, ``Known knowns, known unknowns and unknown unknowns: A 2010 update
  on carotid artery disease,'' {\em Surgeon Journal of the Royal Colleges of
  Surgeons of Edinburgh and Ireland}, vol.~8, no.~2, pp.~79--86, 2010.

\bibitem{dhamija2018reducing}
A.~R. Dhamija, M.~G{\"u}nther, and T.~Boult, ``Reducing network
  agnostophobia,'' in {\em Advances in Neural Information Processing Systems},
  pp.~9157--9168, 2018.

\bibitem{weston2006inference}
J.~Weston, R.~Collobert, F.~Sinz, L.~Bottou, and V.~Vapnik, ``Inference with
  the universum,'' in {\em Proceedings of the 23rd international conference on
  Machine learning}, pp.~1009--1016, ACM, 2006.

\bibitem{maaten2008visualizing}
L.~v.~d. Maaten and G.~Hinton, ``Visualizing data using t-sne,'' {\em Journal
  of machine learning research}, vol.~9, no.~Nov, pp.~2579--2605, 2008.

\bibitem{salakhutdinov2011learning}
R.~Salakhutdinov, A.~Torralba, and J.~Tenenbaum, ``Learning to share visual
  appearance for multiclass object detection,'' in {\em CVPR 2011},
  pp.~1481--1488, IEEE, 2011.

\bibitem{zhu2014capturing}
X.~Zhu, D.~Anguelov, and D.~Ramanan, ``Capturing long-tail distributions of
  object subcategories,'' in {\em Proceedings of the IEEE Conference on
  Computer Vision and Pattern Recognition}, pp.~915--922, 2014.

\bibitem{socher2013zero}
R.~Socher, M.~Ganjoo, C.~D. Manning, and A.~Ng, ``Zero-shot learning through
  cross-modal transfer,'' in {\em Advances in neural information processing
  systems}, pp.~935--943, 2013.

\bibitem{chao2016empirical}
W.-L. Chao, S.~Changpinyo, B.~Gong, and F.~Sha, ``An empirical study and
  analysis of generalized zero-shot learning for object recognition in the
  wild,'' in {\em European Conference on Computer Vision}, pp.~52--68,
  Springer, 2016.

\bibitem{xian2018zero}
Y.~Xian, C.~H. Lampert, B.~Schiele, and Z.~Akata, ``Zero-shot learning-a
  comprehensive evaluation of the good, the bad and the ugly,'' {\em IEEE
  transactions on pattern analysis and machine intelligence}, 2018.

\bibitem{felix2018multi}
R.~Felix, B.~Vijay~Kumar, I.~Reid, and G.~Carneiro, ``Multi-modal
  cycle-consistent generalized zero-shot learning,'' {\em arXiv preprint
  arXiv:1808.00136}, 2018.

\bibitem{gidaris2018dynamic}
S.~Gidaris and N.~Komodakis, ``Dynamic few-shot visual learning without
  forgetting,'' in {\em Proceedings of the IEEE Conference on Computer Vision
  and Pattern Recognition}, pp.~4367--4375, 2018.

\bibitem{chow1970optimum}
C.~Chow, ``On optimum recognition error and reject tradeoff,'' {\em IEEE
  Transactions on information theory}, vol.~16, no.~1, pp.~41--46, 1970.

\bibitem{bartlett2008classification}
P.~L. Bartlett and M.~H. Wegkamp, ``Classification with a reject option using a
  hinge loss,'' {\em Journal of Machine Learning Research}, vol.~9, no.~Aug,
  pp.~1823--1840, 2008.

\bibitem{tax2008growing}
D.~M. Tax and R.~P. Duin, ``Growing a multi-class classifier with a reject
  option,'' {\em Pattern Recognition Letters}, vol.~29, no.~10, pp.~1565--1570,
  2008.

\bibitem{fischer2016optimal}
L.~Fischer, B.~Hammer, and H.~Wersing, ``Optimal local rejection for
  classifiers,'' {\em Neurocomputing}, vol.~214, pp.~445--457, 2016.

\bibitem{fumera2002support}
G.~Fumera and F.~Roli, ``Support vector machines with embedded reject option,''
  in {\em Pattern recognition with support vector machines}, pp.~68--82,
  Springer, 2002.

\bibitem{grandvalet2009support}
Y.~Grandvalet, A.~Rakotomamonjy, J.~Keshet, and S.~Canu, ``Support vector
  machines with a reject option,'' in {\em In Advances in Neural Information
  Processing Systems}, pp.~537--544, 2009.

\bibitem{herbei2006classification}
R.~Herbei and M.~H. Wegkamp, ``Classification with reject option,'' {\em
  Canadian Journal of Statistics}, vol.~34, no.~4, pp.~709--721, 2006.

\bibitem{yuan2010classification}
M.~Yuan and M.~Wegkamp, ``Classification methods with reject option based on
  convex risk minimization,'' {\em Journal of Machine Learning Research},
  vol.~11, no.~Jan, pp.~111--130, 2010.

\bibitem{zhang2006ro}
R.~Zhang and D.~N. Metaxas, ``Ro-svm: Support vector machine with reject option
  for image categorization.,'' in {\em BMVC}, pp.~1209--1218, Citeseer, 2006.

\bibitem{Wegkamp2007Lasso}
M.~Wegkamp, ``Lasso type classifiers with a reject option,'' {\em Electronic
  Journal of Statistics}, vol.~1, no.~3, pp.~155--168, 2007.

\bibitem{Geifman2017Selective}
Y.~Geifman and E.~Y. Ran, ``Selective classification for deep neural
  networks,'' {\em In Advances in Neural Information Processing Systems},
  pp.~4872--4887, 2017.

\bibitem{scholkopf2001estimating}
B.~Sch{\"o}lkopf, J.~C. Platt, J.~Shawe-Taylor, A.~J. Smola, and R.~C.
  Williamson, ``Estimating the support of a high-dimensional distribution,''
  {\em Neural computation}, vol.~13, no.~7, pp.~1443--1471, 2001.

\bibitem{manevitz2001one}
L.~M. Manevitz and M.~Yousef, ``One-class svms for document classification,''
  {\em Journal of machine Learning research}, vol.~2, no.~Dec, pp.~139--154,
  2001.

\bibitem{tax2004support}
D.~M. Tax and R.~P. Duin, ``Support vector data description,'' {\em Machine
  learning}, vol.~54, no.~1, pp.~45--66, 2004.

\bibitem{khan2009survey}
S.~S. Khan and M.~G. Madden, ``A survey of recent trends in one class
  classification,'' in {\em Irish conference on artificial intelligence and
  cognitive science}, pp.~188--197, Springer, 2009.

\bibitem{Jin2004Face}
H.~Jin, Q.~Liu, and H.~Lu, ``Face detection using one-class-based support
  vectors,'' in {\em IEEE International Conference on Automatic Face and
  Gesture Recognition}, pp.~457--462, 2004.

\bibitem{Wu2009A}
M.~Wu and J.~Ye, ``A small sphere and large margin approach for novelty
  detection using training data with outliers,'' {\em IEEE Transactions on
  Pattern Analysis and Machine Intelligence}, vol.~31, no.~11, pp.~2088--2092,
  2009.

\bibitem{Cevikalp2012Efficient}
H.~Cevikalp and B.~Triggs, ``Efficient object detection using cascades of
  nearest convex model classifiers,'' in {\em Computer Vision and Pattern
  Recognition}, pp.~3138--3145, 2012.

\bibitem{pimentel2014review}
M.~A. Pimentel, D.~A. Clifton, L.~Clifton, and L.~Tarassenko, ``A review of
  novelty detection,'' {\em Signal Processing}, vol.~99, pp.~215--249, 2014.

\bibitem{gornitz2018support}
N.~G{\"o}rnitz, L.~A. Lima, K.-R. M{\"u}ller, M.~Kloft, and S.~Nakajima,
  ``Support vector data descriptions and $ k $-means clustering: One class?,''
  {\em IEEE Transactions on Neural Networks and Learning Systems}, vol.~29,
  no.~9, pp.~3994--4006, 2018.

\bibitem{bodesheim2013kernel}
P.~Bodesheim, A.~Freytag, E.~Rodner, M.~Kemmler, and J.~Denzler, ``Kernel null
  space methods for novelty detection,'' in {\em Proceedings of the IEEE
  Conference on Computer Vision and Pattern Recognition}, pp.~3374--3381, 2013.

\bibitem{Phillips2005Evaluation}
P.~J. Phillips, P.~Grother, and R.~Micheals, {\em Evaluation Methods in Face
  Recognition}.
\newblock Handbook of Face Recognition. Springer New York, 2005.

\bibitem{Li2005Open}
F.~Li and H.~Wechsler, ``Open set face recognition using transduction,'' {\em
  IEEE Transactions on Pattern Analysis and Machine Intelligence}, vol.~27,
  no.~11, pp.~1686--1697, 2005.

\bibitem{wu2007novel}
Q.~Wu, C.~Jia, and W.~Chen, ``A novel classification-rejection sphere svms for
  multi-class classification problems,'' in {\em Natural Computation, 2007.
  ICNC 2007. Third International Conference on}, vol.~1, pp.~34--38, IEEE,
  2007.

\bibitem{wang2009support}
Y.-C.~F. Wang and D.~Casasent, ``A support vector hierarchical method for
  multi-class classification and rejection,'' {\em In International Joint
  Conference on Neural Networks}, pp.~3281--3288, 2009.

\bibitem{Heflin2012Detecting}
B.~Heflin, W.~Scheirer, and T.~E. Boult, ``Detecting and classifying scars,
  marks, and tattoos found in the wild,'' in {\em IEEE Fifth International
  Conference on Biometrics: Theory, Applications and Systems}, pp.~31--38,
  2012.

\bibitem{Pritsos2013Open}
D.~A. Pritsos and E.~Stamatatos, ``Open-set classification for automated genre
  identification,'' in {\em European Conference on Advances in Information
  Retrieval}, pp.~207--217, 2013.

\bibitem{Cevikalp2013Face}
H.~Cevikalp, B.~Triggs, and V.~Franc, ``Face and landmark detection by using
  cascade of classifiers,'' in {\em IEEE International Conference and Workshops
  on Automatic Face and Gesture Recognition}, pp.~1--7, 2013.

\bibitem{Cevikalp2017Best}
H.~Cevikalp, ``Best fitting hyperplanes for classification,'' {\em IEEE
  Transactions on Pattern Analysis and Machine Intelligence}, vol.~39, no.~6,
  p.~1076, 2017.

\bibitem{Scherreik2016Open}
M.~D. Scherreik and B.~D. Rigling, ``Open set recognition for automatic target
  classification with rejection,'' {\em IEEE Transactions on Aerospace and
  Electronic Systems}, vol.~52, no.~2, pp.~632--642, 2016.

\bibitem{Cevikalp2017Polyhedral}
H.~Cevikalp and B.~Triggs, ``Polyhedral conic classifiers for visual object
  detection and classification,'' in {\em IEEE Conference on Computer Vision
  and Pattern Recognition}, pp.~4114--4122, 2017.

\bibitem{Cevikalp2017Fast}
H.~Cevikalp and H.~S. Yavuz, ``Fast and accurate face recognition with image
  sets,'' in {\em IEEE International Conference on Computer Vision Workshop},
  pp.~1564--1572, 2017.

\bibitem{Zhang2017Sparse}
H.~Zhang and V.~Patel, ``Sparse representation-based open set recognition,''
  {\em IEEE Transactions on Pattern Analysis and Machine Intelligence},
  vol.~39, no.~8, pp.~1690--1696, 2017.

\bibitem{Bendale2015Towards}
A.~Bendale and T.~Boult, ``Towards open world recognition,'' in {\em in
  Proceedings of the IEEE Conference on Computer Vision and Pattern
  Recognition}, pp.~1893--1902, 2015.

\bibitem{J2017Nearest}
P.~R.~M. Júnior, R.~M.~D. Souza, R.~D.~O. Werneck, B.~V. Stein, D.~V.
  Pazinato, W.~R.~D. Almeida, O.~A.~B. Penatti, R.~D.~S. Torres, and A.~Rocha,
  ``Nearest neighbors distance ratio open-set classifier,'' {\em Machine
  Learning}, vol.~106, no.~3, pp.~359--386, 2017.

\bibitem{Rudd2018The}
E.~M. Rudd, L.~P. Jain, W.~J. Scheirer, and T.~E. Boult, ``The extreme value
  machine,'' {\em IEEE Transactions on Pattern Analysis and Machine
  Intelligence}, vol.~40, no.~3, pp.~762--768, 2018.

\bibitem{Vignotto2018Extreme}
E.~Vignotto and S.~Engelke, ``Extreme value theory for open set
  classification-gpd and gev classifiers,'' {\em arXiv preprint
  arXiv:1808.09902}, 2018.

\bibitem{Fei2016Breaking}
G.~Fei and B.~Liu, ``Breaking the closed world assumption in text
  classification,'' in {\em Conference of the North American Chapter of the
  Association for Computational Linguistics: Human Language Technologies},
  pp.~506--514, 2016.

\bibitem{Vareto2018Towards}
R.~Vareto, S.~Silva, F.~Costa, and W.~R. Schwartz, ``Towards open-set face
  recognition using hashing functions,'' in {\em IEEE International Joint
  Conference on Biometrics}, 2018.

\bibitem{Neira2018Data}
M.~A.~C. Neira, P.~R.~M. Junior, A.~Rocha, and R.~D.~S. Torres, ``Data-fusion
  techniques for open-set recognition problems,'' {\em IEEE Access}, vol.~6,
  pp.~21242--21265, 2018.

\bibitem{Bendale2015Towards1}
A.~Bendale and T.~E. Boult, ``Towards open set deep networks,'' {\em
  Proceedings of the IEEE conference on computer vision and pattern
  recognition}, pp.~1563--1572, 2016.

\bibitem{Rozsa2017Adversarial}
A.~Rozsa, M.~Günther, and T.~E. Boult, ``Adversarial robustness: Softmax
  versus openmax,'' {\em arXiv preprint arXiv:1708.01697}, 2017.

\bibitem{Hassen2018Learning}
M.~Hassen and P.~K. Chan, ``Learning a neural-network-based representation for
  open set recognition,'' {\em arXiv preprint arXiv:1802.04365}, 2018.

\bibitem{prakhya2017open}
S.~Prakhya, V.~Venkataram, and J.~Kalita, ``Open set text classification using
  convolutional neural networks,'' in {\em International Conference on Natural
  Language Processing}, 2017.

\bibitem{Shu2017DOC}
L.~Shu, H.~Xu, and B.~Liu, ``Doc: Deep open classification of text documents,''
  {\em arXiv preprint arXiv:1709.08716}, 2017.

\bibitem{kardan2017mitigating}
N.~Kardan and K.~O. Stanley, ``Mitigating fooling with competitive overcomplete
  output layer neural networks,'' in {\em 2017 International Joint Conference
  on Neural Networks (IJCNN)}, pp.~518--525, IEEE, 2017.

\bibitem{Cardoso2015A}
D.~O. Cardoso, F.~Franca, and J.~Gama, ``A bounded neural network for open set
  recognition,'' in {\em International Joint Conference on Neural Networks},
  pp.~1--7, 2015.

\bibitem{Cardoso2017Weightless}
D.~O. Cardoso, J.~Gama, and F.~M.~G. França, ``Weightless neural networks for
  open set recognition,'' {\em Machine Learning}, no.~106(9-10),
  pp.~1547--1567, 2017.

\bibitem{yoshihashi2019classification}
R.~Yoshihashi, W.~Shao, R.~Kawakami, S.~You, M.~Iida, and T.~Naemura,
  ``Classification-reconstruction learning for open-set recognition,'' in {\em
  Proceedings of the IEEE Conference on Computer Vision and Pattern
  Recognition}, pp.~4016--4025, 2019.

\bibitem{Shu2018Unseen}
L.~Shu, H.~Xu, and B.~Liu, ``Unseen class discovery in open-world
  classification,'' {\em arXiv preprint arXiv:1801.05609}, 2018.

\bibitem{oza2019c2ae}
P.~Oza and V.~M. Patel, ``C2ae: Class conditioned auto-encoder for open-set
  recognition,'' {\em arXiv preprint arXiv:1904.01198}, 2019.

\bibitem{Ge2017Generative}
Z.~Y. Ge, S.~Demyanov, Z.~Chen, and R.~Garnavi, ``Generative openmax for
  multi-class open set classification,'' {\em arXiv preprint arXiv:1707.07418},
  2017.

\bibitem{Neal2018Open}
L.~Neal, M.~Olson, X.~Fern, W.~Wong, and F.~Li, ``Open set learning with
  counterfactual images,'' {\em Proceedings of the European Conference on
  Computer Vision (ECCV)}, pp.~613--628, 2018.

\bibitem{Jo2018Open}
I.~Jo, J.~Kim, H.~Kang, Y.-D. Kim, and S.~Choi, ``Open set recognition by
  regularising classifier with fake data generated by generative adversarial
  networks,'' {\em IEEE International Conference on Acoustics, Speech and
  Signal Processing}, pp.~2686--2690, 2018.

\bibitem{Yu2017Open}
Y.~Yu, W.-Y. Qu, N.~Li, and Z.~Guo, ``Open-category classification by
  adversarial sample generation,'' {\em arXiv preprint arXiv:1705.08722}, 2017.

\bibitem{yang2019open}
Y.~Yang, C.~Hou, Y.~Lang, D.~Guan, D.~Huang, and J.~Xu, ``Open-set human
  activity recognition based on micro-doppler signatures,'' {\em Pattern
  Recognition}, vol.~85, pp.~60--69, 2019.

\bibitem{Geng2018Collective}
C.~Geng and S.~Chen, ``Collective decision for open set recognition,'' {\em
  arXiv preprint arXiv:1806.11258v4}, 2018.

\bibitem{Bouchard2004The}
G.~Bouchard and B.~Triggs, ``The trade-off between generative and
  discriminative classifiers,'' {\em Proceedings in Computational Statistics
  Symposium of Iasc}, pp.~721--728, 2004.

\bibitem{Lasserre2006Principled}
J.~A. Lasserre, C.~M. Bishop, and T.~P. Minka, ``Principled hybrids of
  generative and discriminative models,'' in {\em Computer Vision and Pattern
  Recognition, 2006 IEEE Computer Society Conference on}, pp.~87--94, 2006.

\bibitem{Cortes1995Support}
C.~Cortes and V.~Vapnik, ``Support-vector networks,'' in {\em Machine
  Learning}, pp.~273--297, 1995.

\bibitem{kotz2000extreme}
S.~Kotz and S.~Nadarajah, {\em Extreme value distributions: theory and
  applications}.
\newblock World Scientific, 2000.

\bibitem{Cruz2017Open}
S.~Cruz, C.~Coleman, E.~M. Rudd, and T.~E. Boult, ``Open set intrusion
  recognition for fine-grained attack categorization,'' in {\em IEEE
  International Symposium on Technologies for Homeland Security}, pp.~1--6,
  2017.

\bibitem{Rudd2017A}
E.~Rudd, A.~Rozsa, M.~Gunther, and T.~Boult, ``A survey of stealth malware:
  Attacks, mitigation measures, and steps toward autonomous open world
  solutions,'' {\em IEEE Communications Surveys and Tutorials}, vol.~19, no.~2,
  pp.~1145--1172, 2017.

\bibitem{Rafail2006Separation}
R.~N. Gasimov and G.~Ozturk, ``Separation via polyhedral conic functions,''
  {\em Optimization Methods and Software}, vol.~21, no.~4, pp.~527--540, 2006.

\bibitem{Wright2010Sparse}
J.~Wright, Y.~Ma, J.~Mairal, G.~Sapiro, T.~S. Huang, and S.~Yan, ``Sparse
  representation for computer vision and pattern recognition,'' {\em
  Proceedings of the IEEE}, vol.~98, no.~6, pp.~1031--1044, 2010.

\bibitem{Rubinstein2010Dictionaries}
R.~Rubinstein, A.~M. Bruckstein, and M.~Elad, ``Dictionaries for sparse
  representation modeling,'' {\em Proceedings of the IEEE}, vol.~98, no.~6,
  pp.~1045--1057, 2010.

\bibitem{peng2018maximum}
J.~Peng, L.~Li, and Y.~Y. Tang, ``Maximum likelihood estimation-based joint
  sparse representation for the classification of hyperspectral remote sensing
  images,'' {\em IEEE Transactions on Neural Networks and Learning Systems},
  2018.

\bibitem{Wright2009Robust}
J.~Wright, A.~Y. Yang, A.~Ganesh, S.~S. Sastry, and Y.~Ma, ``Robust face
  recognition via sparse representation,'' {\em IEEE Transactions on Pattern
  Analysis and Machine Intelligence}, vol.~31, no.~2, pp.~210--227, 2009.

\bibitem{Mensink2013Distance}
T.~Mensink, J.~Verbeek, F.~Perronnin, and G.~Csurka, ``Distance-based image
  classification: Generalizing to new classes at near-zero cost,'' {\em IEEE
  transactions on pattern analysis and machine intelligence}, vol.~35, no.~11,
  pp.~2624--2637, 2013.

\bibitem{Ristin2014Incremental}
M.~Ristin, M.~Guillaumin, J.~Gall, and L.~V. Gool, ``Incremental learning of
  ncm forests for large-scale image classification,'' {\em Computer Vision and
  Pattern Recognition}, pp.~3654--3661, 2014.

\bibitem{Papa2009Supervised}
J.~P. Papa and C.~T.~N. Suzuki, {\em Supervised pattern classification based on
  optimum-path forest}.
\newblock John Wiley and Sons, Inc., 2009.

\bibitem{garg2003margin}
A.~Garg and D.~Roth, ``Margin distribution and learning,'' in {\em Proceedings
  of the 20th International Conference on Machine Learning (ICML-03)},
  pp.~210--217, 2003.

\bibitem{reyzin2006boosting}
L.~Reyzin and R.~E. Schapire, ``How boosting the margin can also boost
  classifier complexity,'' in {\em Proceedings of the 23rd international
  conference on Machine learning}, pp.~753--760, ACM, 2006.

\bibitem{aiolli2008kernel}
F.~Aiolli, G.~Da~San~Martino, and A.~Sperduti, ``A kernel method for the
  optimization of the margin distribution,'' in {\em International Conference
  on Artificial Neural Networks}, pp.~305--314, Springer, 2008.

\bibitem{pelckmans2008risk}
K.~Pelckmans, J.~Suykens, and B.~D. Moor, ``A risk minimization principle for a
  class of parzen estimators,'' in {\em Advances in Neural Information
  Processing Systems}, pp.~1137--1144, 2008.

\bibitem{G2017Toward}
M.~Günther, S.~Cruz, E.~M. Rudd, and T.~E. Boult, ``Toward open-set face
  recognition,'' {\em Conference on Computer Vision and Pattern Recognition
  (CVPR) Workshops}, 2017.

\bibitem{henrydoss2017incremental}
J.~Henrydoss, S.~Cruz, E.~M. Rudd, T.~E. Boult, {\em et~al.}, ``Incremental
  open set intrusion recognition using extreme value machine,'' in {\em Machine
  Learning and Applications (ICMLA), 2017 16th IEEE International Conference
  on}, pp.~1089--1093, IEEE, 2017.

\bibitem{Nguyen2014Deep}
A.~Nguyen, J.~Yosinski, and J.~Clune, ``Deep neural networks are easily fooled:
  High confidence predictions for unrecognizable images,'' {\em Proceedings of
  the IEEE Conference on Computer Vision and Pattern Recognition},
  pp.~427--436, 2015.

\bibitem{Goodfellow2015Explaining}
I.~Goodfellow, J.~Shelns, and C.~Szegedy, ``Explaining and harnessing
  adversarial examples,'' {\em In International Conference on Learning
  Representations. Computational and Biological Learning Society}, 2015.

\bibitem{Szegedy2014Intriguing}
C.~Szegedy, W.~Zaremba, I.~Sutskever, J.~Bruna, D.~Erhan, I.~Goodfellow, and
  R.~Fergus, ``Intriguing properties of neural networks,'' {\em In
  International Conference on Learning Representations. Computational and
  Biological Learning Society}, 2014.

\bibitem{da2014learning}
Q.~Da, Y.~Yu, and Z.-H. Zhou, ``Learning with augmented class by exploiting
  unlabeled data,'' in {\em Twenty-Eighth AAAI Conference on Artificial
  Intelligence}, 2014.

\bibitem{masana2018metric}
M.~Masana, I.~Ruiz, J.~Serrat, J.~van~de Weijer, and A.~M. Lopez, ``Metric
  learning for novelty and anomaly detection,'' {\em arXiv preprint
  arXiv:1808.05492}, 2018.

\bibitem{reed2001pareto}
W.~J. Reed, ``The pareto, zipf and other power laws,'' {\em Economics letters},
  vol.~74, no.~1, pp.~15--19, 2001.

\bibitem{liu2019large}
Z.~Liu, Z.~Miao, X.~Zhan, J.~Wang, B.~Gong, and S.~X. Yu, ``Large-scale
  long-tailed recognition in an open world,'' in {\em Proceedings of the IEEE
  Conference on Computer Vision and Pattern Recognition}, pp.~2537--2546, 2019.

\bibitem{goodfellow2016nips}
I.~Goodfellow, ``Nips 2016 tutorial: Generative adversarial networks,'' {\em
  arXiv preprint arXiv:1701.00160}, 2016.

\bibitem{teh2011dirichlet}
Y.~W. Teh, ``Dirichlet process,'' in {\em Encyclopedia of machine learning},
  pp.~280--287, Springer, 2011.

\bibitem{Gershman2012A}
S.~J. Gershman and D.~M. Blei, ``A tutorial on bayesian nonparametric models,''
  {\em Journal of Mathematical Psychology}, vol.~56, no.~1, pp.~1--12, 2012.

\bibitem{fan2013online}
W.~Fan and N.~Bouguila, ``Online learning of a dirichlet process mixture of
  beta-liouville distributions via variational inference,'' {\em IEEE
  Transactions on Neural Networks and Learning Systems}, vol.~24, no.~11,
  pp.~1850--1862, 2013.

\bibitem{Teh2006Hierarchical}
Y.~W. Teh, M.~I. Jordan, M.~J. Beal, and D.~M. Blei, ``Hierarchical dirichlet
  processes,'' {\em Publications of the American Statistical Association},
  vol.~101, no.~476, pp.~1566--1581, 2006.

\bibitem{bargi2018adon}
A.~Bargi, R.~Y. Da~Xu, and M.~Piccardi, ``Adon hdp-hmm: an adaptive online
  model for segmentation and classification of sequential data,'' {\em IEEE
  Transactions on Neural Networks and Learning Systems}, vol.~29, no.~9,
  pp.~3953--3968, 2018.

\bibitem{cauwenberghs2001incremental}
G.~Cauwenberghs and T.~Poggio, ``Incremental and decremental support vector
  machine learning,'' in {\em Advances in neural information processing
  systems}, pp.~409--415, 2001.

\bibitem{crammer2006online}
K.~Crammer, O.~Dekel, J.~Keshet, S.~Shalev-Shwartz, and Y.~Singer, ``Online
  passive-aggressive algorithms,'' {\em Journal of Machine Learning Research},
  vol.~7, no.~Mar, pp.~551--585, 2006.

\bibitem{fink2006online}
M.~Fink, S.~Shalev-Shwartz, Y.~Singer, and S.~Ullman, ``Online multiclass
  learning by interclass hypothesis sharing,'' in {\em Proceedings of the 23rd
  international conference on Machine learning}, pp.~313--320, ACM, 2006.

\bibitem{yeh2008dynamic}
T.~Yeh and T.~Darrell, ``Dynamic visual category learning,'' in {\em 2008 IEEE
  Conference on Computer Vision and Pattern Recognition}, pp.~1--8, IEEE, 2008.

\bibitem{muhlbaier2008learn}
M.~D. Muhlbaier, A.~Topalis, and R.~Polikar, ``Learn++ nc: Combining ensemble
  of classifiers with dynamically weighted consult-and-vote for efficient
  incremental learning of new classes,'' {\em IEEE transactions on neural
  networks}, vol.~20, no.~1, pp.~152--168, 2008.

\bibitem{kuzborskij2013n}
I.~Kuzborskij, F.~Orabona, and B.~Caputo, ``From n to n+ 1: Multiclass transfer
  incremental learning,'' in {\em Proceedings of the IEEE Conference on
  Computer Vision and Pattern Recognition}, pp.~3358--3365, 2013.

\bibitem{de2016online}
R.~De~Rosa, T.~Mensink, and B.~Caputo, ``Online open world recognition,'' {\em
  arXiv preprint arXiv:1604.02275}, 2016.

\bibitem{doan2017overcoming}
T.~Doan and J.~Kalita, ``Overcoming the challenge for text classification in
  the open world,'' in {\em Computing and Communication Workshop and Conference
  (CCWC), 2017 IEEE 7th Annual}, pp.~1--7, IEEE, 2017.

\bibitem{Lonij2017Open}
V.~P.~A. Lonij, A.~Rawat, and M.~I. Nicolae, ``Open-world visual recognition
  using knowledge graphs,'' {\em arXiv preprint arXiv:1708.08310}, 2017.

\bibitem{shi2018odn}
Y.~Shi, Y.~Wang, Y.~Zou, Q.~Yuan, Y.~Tian, and Y.~Shu, ``Odn: Opening the deep
  network for open-set action recognition,'' in {\em 2018 IEEE International
  Conference on Multimedia and Expo (ICME)}, pp.~1--6, IEEE, 2018.

\bibitem{Xu2018Learning}
H.~Xu, B.~Liu, and P.~S. Yu, ``Learning to accept new classes without
  training,'' {\em arXiv preprint arXiv:1809.06004}, 2018.

\bibitem{frey1991letter}
P.~W. Frey and D.~J. Slate, ``Letter recognition using holland-style adaptive
  classifiers,'' {\em Machine learning}, vol.~6, no.~2, pp.~161--182, 1991.

\bibitem{bilenko2004integrating}
M.~Bilenko, S.~Basu, and R.~J. Mooney, ``Integrating constraints and metric
  learning in semi-supervised clustering,'' in {\em Proceedings of the
  twenty-first international conference on Machine learning}, p.~11, ACM, 2004.

\bibitem{nene1996columbia}
S.~A. Nene, S.~K. Nayar, H.~Murase, {\em et~al.}, ``Columbia object image
  library (coil-20),'' 1996.

\bibitem{georghiades2001few}
A.~S. Georghiades, P.~N. Belhumeur, and D.~J. Kriegman, ``From few to many:
  Illumination cone models for face recognition under variable lighting and
  pose,'' {\em IEEE Transactions on Pattern Analysis \& Machine Intelligence},
  no.~6, pp.~643--660, 2001.

\bibitem{lecun1998gradient}
Y.~LeCun, L.~Bottou, Y.~Bengio, P.~Haffner, {\em et~al.}, ``Gradient-based
  learning applied to document recognition,'' {\em Proceedings of the IEEE},
  vol.~86, no.~11, pp.~2278--2324, 1998.

\bibitem{netzer2011reading}
Y.~Netzer, T.~Wang, A.~Coates, A.~Bissacco, B.~Wu, and A.~Y. Ng, ``Reading
  digits in natural images with unsupervised feature learning,'' {\em In: NIPS
  workshop on deep learning and unsupervised feature learning}, 2011.

\bibitem{krizhevsky2009learning}
A.~Krizhevsky, G.~Hinton, {\em et~al.}, ``Learning multiple layers of features
  from tiny images,'' tech. rep., Citeseer, 2009.

\bibitem{le2015tiny}
Y.~Le and X.~Yang, ``Tiny imagenet visual recognition challenge,'' {\em CS
  231N}, 2015.

\bibitem{russakovsky2015imagenet}
O.~Russakovsky, J.~Deng, H.~Su, J.~Krause, S.~Satheesh, S.~Ma, Z.~Huang,
  A.~Karpathy, A.~Khosla, M.~Bernstein, {\em et~al.}, ``Imagenet large scale
  visual recognition challenge,'' {\em International journal of computer
  vision}, vol.~115, no.~3, pp.~211--252, 2015.

\bibitem{Anna2009A}
C.~Anna, L.~Günter, R.~H. Joachim, and M.~Klaus, ``A systematic analysis of
  performance measures for classification tasks,'' {\em Information Processing
  and Management}, vol.~45, no.~4, pp.~427--437, 2009.

\bibitem{Youden1950Index}
W.~J. Youden, ``Index for rating diagnostic tests.,'' {\em Cancer}, vol.~3,
  no.~1, pp.~32--35, 1950.

\bibitem{Sokolova2006Beyond}
M.~Sokolova, N.~Japkowicz, and S.~Szpakowicz, {\em Beyond Accuracy, F-Score and
  ROC: A Family of Discriminant Measures for Performance Evaluation}.
\newblock Springer Berlin Heidelberg, 2006.

\bibitem{wang2018deep}
M.~Wang and W.~Deng, ``Deep visual domain adaptation: A survey,'' {\em
  Neurocomputing}, vol.~312, pp.~135--153, 2018.

\bibitem{Cai2010A}
W.~Cai, S.~Chen, and D.~Zhang, ``A multiobjective simultaneous learning
  framework for clustering and classification,'' {\em IEEE Transactions on
  Neural Networks}, vol.~21, no.~2, pp.~185--200, 2010.

\bibitem{Qian2012Simultaneous}
Q.~Qian, S.~Chen, and W.~Cai, ``Simultaneous clustering and classification over
  cluster structure representation,'' {\em Pattern Recognition}, vol.~45,
  no.~6, pp.~2227--2236, 2012.

\bibitem{Mu2017Classification}
X.~Mu, M.~T. Kai, and Z.~H. Zhou, ``Classification under streaming emerging new
  classes: A solution using completely-random trees,'' {\em IEEE Transactions
  on Knowledge and Data Engineering}, vol.~29, no.~8, pp.~1605--1618, 2017.

\bibitem{Liu2018Open}
S.~Liu, R.~Garrepalli, T.~G. Dietterich, A.~Fern, and D.~Hendrycks, ``Open
  category detection with pac guarantees,'' {\em arXiv preprint
  arXiv:1808.00529}, 2018.

\bibitem{Vyas2018Out}
A.~Vyas, N.~Jammalamadaka, X.~Zhu, D.~Das, B.~Kaul, and T.~L. Willke,
  ``Out-of-distribution detection using an ensemble of self supervised
  leave-out classifiers,'' {\em arXiv preprint arXiv:1809.03576}, 2018.

\bibitem{dong2018learning}
H.~Dong, Y.~Fu, L.~Sigal, S.~J. Hwang, Y.-G. Jiang, and X.~Xue, ``Learning to
  separate domains in generalized zero-shot and open set learning: a
  probabilistic perspective,'' {\em arXiv preprint arXiv:1810.07368}, 2018.

\bibitem{wang2017joint}
L.~Wang and S.~Chen, ``Joint representation classification for collective face
  recognition,'' {\em Pattern Recognition}, vol.~63, pp.~182--192, 2017.

\bibitem{Busto2017Open}
P.~P. Busto and J.~Gall, ``Open set domain adaptation,'' in {\em IEEE
  International Conference on Computer Vision}, pp.~754--763, 2017.

\bibitem{Saito2018Open}
K.~Saito, S.~Yamamoto, Y.~Ushiku, and T.~Harada, ``Open set domain adaptation
  by backpropagation,'' {\em arXiv preprint arXiv:1804.10427}, 2018.

\bibitem{Baktashmotlagh2018Learning}
M.~Baktashmotlagh, M.~Faraki, T.~Drummond, and M.~Salzmann, ``Learning
  factorized representations for open-set domain adaptation,'' {\em arXiv
  preprint arXiv:1805.12277}, 2018.

\bibitem{pham2018bayesian}
T.~Pham, B.~Vijay~Kumar, T.-T. Do, G.~Carneiro, and I.~Reid, ``Bayesian
  semantic instance segmentation in open set world,'' {\em arXiv preprint
  arXiv:1806.00911}, 2018.

\bibitem{Liu2019Active}
Z.-Y. Liu and S.-J. Huang, ``Active sampling for open-set classification
  without initial annotation,'' {\em In: Proceedings of the 33nd AAAI
  Conference on Artificial Intelligence}, 2019.

\bibitem{lee2018multi}
C.-W. Lee, W.~Fang, C.-K. Yeh, and Y.-C. Frank~Wang, ``Multi-label zero-shot
  learning with structured knowledge graphs,'' in {\em Proceedings of the IEEE
  Conference on Computer Vision and Pattern Recognition}, pp.~1576--1585, 2018.

\bibitem{Fu2016Semi}
Y.~Fu and L.~Sigal, ``Semi-supervised vocabulary-informed learning,'' in {\em
  Computer Vision and Pattern Recognition}, pp.~5337--5346, 2016.

\bibitem{Fu2017Vocabulary}
Y.~Fu, H.~Z. Dong, Y.~F. Ma, Z.~Zhang, and X.~Xue, ``Vocabulary-informed
  extreme value learning,'' {\em arXiv preprint arXiv:1705.09887}, 2017.

\bibitem{guadarrama2014open}
S.~Guadarrama, E.~Rodner, K.~Saenko, N.~Zhang, R.~Farrell, J.~Donahue, and
  T.~Darrell, ``Open-vocabulary object retrieval.,'' in {\em Robotics: science
  and systems}, vol.~2, p.~6, Citeseer, 2014.

\bibitem{guadarrama2016understanding}
S.~Guadarrama, E.~Rodner, K.~Saenko, and T.~Darrell, ``Understanding object
  descriptions in robotics by open-vocabulary object retrieval and detection,''
  {\em The International Journal of Robotics Research}, vol.~35, no.~1-3,
  pp.~265--280, 2016.

\bibitem{zheng2016towards}
W.-S. Zheng, S.~Gong, and T.~Xiang, ``Towards open-world person
  re-identification by one-shot group-based verification,'' {\em IEEE
  Transactions on Pattern Analysis and Machine Intelligence}, vol.~38, no.~3,
  pp.~591--606, 2016.

\bibitem{sattar2015prediction}
H.~Sattar, S.~Muller, M.~Fritz, and A.~Bulling, ``Prediction of search targets
  from fixations in open-world settings,'' in {\em Proceedings of the IEEE
  Conference on Computer Vision and Pattern Recognition}, pp.~981--990, 2015.

\bibitem{Zhao2017Open}
H.~Zhao, X.~Puig, B.~Zhou, S.~Fidler, and A.~Torralba, ``Open vocabulary scene
  parsing,'' in {\em Proc. IEEE Conf. Computer Vision and Pattern Recognition},
  2017.

\bibitem{Cherkassky2011Practical}
V.~Cherkassky, S.~Dhar, and W.~Dai, ``Practical conditions for effectiveness of
  the universum learning,'' {\em IEEE Transactions on Neural Networks},
  vol.~22, no.~8, pp.~1241--55, 2011.

\bibitem{Qi2012Twin}
Z.~Qi, Y.~Tian, and Y.~Shi, ``Twin support vector machine with universum
  data,'' {\em Neural Networks}, vol.~36, no.~3, pp.~112--119, 2012.

\bibitem{lee2018hierarchical}
K.~Lee, K.~Lee, K.~Min, Y.~Zhang, J.~Shin, and H.~Lee, ``Hierarchical novelty
  detection for visual object recognition,'' in {\em Proceedings of the IEEE
  Conference on Computer Vision and Pattern Recognition}, pp.~1034--1042, 2018.

\bibitem{Bendavid2010A}
S.~Bendavid, J.~Blitzer, K.~Crammer, A.~Kulesza, F.~Pereira, and J.~W. Vaughan,
  ``A theory of learning from different domains,'' {\em Machine Learning},
  vol.~79, no.~1-2, pp.~151--175, 2010.

\bibitem{Sun2015A}
S.~Sun, H.~Shi, and Y.~Wu, {\em A survey of multi-source domain adaptation}.
\newblock Elsevier Science Publishers B. V., 2015.

\bibitem{Liu2017Structure}
H.~Liu, M.~Shao, and Y.~Fu, ``Structure-preserved multi-source domain
  adaptation,'' in {\em IEEE International Conference on Data Mining},
  pp.~1059--1064, 2017.

\bibitem{Yu2018Unsupervised}
H.~Yu, M.~Hu, and S.~Chen, ``Unsupervised domain adaptation without exactly
  shared categories,'' {\em arXiv preprint arXiv:1809.00852}, 2018.

\bibitem{dietterich2017steps}
T.~G. Dietterich, ``Steps toward robust artificial intelligence,'' {\em AI
  Magazine}, vol.~38, no.~3, pp.~3--24, 2017.

\end{thebibliography}

\end{document}